%% file: main_arxiv.tex
\documentclass[11pt]{article}

\usepackage[margin=1in]{geometry}
\usepackage{authblk}

\usepackage{float}
\usepackage[table]{xcolor}
\usepackage{graphicx,enumitem,amsmath,amssymb,ulem,subcaption,comment}
\usepackage{algorithm,algpseudocode}
\usepackage{placeins,tabularx,booktabs,multirow}
\usepackage[authoryear,sort]{natbib}
\usepackage{hyperref}
\usepackage{colortbl}
\usepackage{url}
\usepackage{setspace}

\hypersetup{
    colorlinks=true,
    linkcolor=blue,
    citecolor=blue,
    urlcolor=blue,
    filecolor=blue,
    pdfborder={0 0 0}
}

\title{Deep Reinforcement Learning for Dynamic Battery Management of Autonomous Order Pickers}

\author[1]{Taniya Shaji}
\author[1]{Abhay Sobhanan\thanks{Corresponding author. Email: \texttt{abhay.sobhanan@iimb.ac.in}}}
\author[2]{Christof Defryn}

\affil[1]{Indian Institute of Management Bangalore, Bannerghatta Road, Bengaluru 560076, Karnataka, India}
\affil[2]{University of Antwerp, Prinsstraat 13, Antwerp 2000, Belgium}

\date{}

\begin{document}
\maketitle

\onehalfspacing 

\begin{abstract}
Battery charging of Autonomous Mobile Robots (AMRs) in warehouses is a critical operational challenge that heavily impacts both order processing times and throughput. In this study, we address the dynamic AMR charging problem under stochastic order arrivals, where robots must learn optimal charging decisions. Traditional fixed-rule heuristics often prove suboptimal in dynamic environments and fail to account for multi-AMR coordination, leading to severe resource inefficiencies. To overcome these limitations, we propose a Proximal Policy Optimization (PPO)-based Deep Reinforcement Learning (DRL) framework designed for multi-block warehouses with fixed charging stations. Our model dynamically learns two key decisions: charging station selection and optimal charging duration, explicitly accounting for anticipated queuing times at the stations. Extensive numerical experiments benchmark the proposed model against state-of-the-art DRL and traditional heuristic approaches.
Results demonstrate that our PPO framework increases order-completion rates by up to 6\% compared to the strongest baseline, while significantly reducing the total time dedicated to recharging operations.
Furthermore, we validate the model's robustness across diverse warehouse configurations and stochastic arrival rates. Finally, we interpret the learned DRL policy, offering valuable operational insights into its superiority over standard benchmarks.
\end{abstract}



\input{sections/1_introduction}

\input{sections/2_literature_review}
\input{sections/3_problem_description}
\input{sections/4_mdp_formulation}

\input{sections/5_methodology_PPO}

\input{sections/6_results_PPO}
\input{sections/8_benchmarking_PPO}
\input{sections/7_explainability}

\input{sections/9_sensitivity}


\section{Conclusions}
With the increasing dependence on automatic operation of warehouses and AMRs, this study investigates optimal charging strategies for AMRs operating in a multi-block warehouse. A PPO-based deep reinforcement learning framework is proposed to learn charging strategies instead of relying on fixed rules, which are currently dominant in the industry. The key decisions learned include deciding when to start and stop charging, which charging station to use to avoid congestion, and when to go to the depot, under dynamically arriving orders while continuously accounting for battery state and agent location of the AMRs in real time. The framework was evaluated under multiple environment configurations and under stochastic demands. Across all scenarios, the proposed approach achieved the highest order completion rate, lowest waiting time at charging stations and the lowest time spent charging per order. The results demonstrate the effectiveness and robustness of the proposed framework for dynamic battery management, thereby reducing energy consumption and operational costs.

Based on the interpretability analysis, we propose practical decision rules for managing AMRs in operational warehouses:
\begin{itemize}[itemsep=0.0pt, topsep=0.0pt]
  \item Direct AMRs to the nearest charging stations, and strategically position these stations to ensure uniform coverage across all agents.
  \item Position charging stations to grant easier access to agents operating farthest from the depot.
  \item Terminate charging cycles before reaching full capacity if the battery levels of nearby agents, or those sharing the same station, fall below a critical threshold.
  \item Initiate depot visits after an agent's carrying capacity has been fully utilized.
\end{itemize}

To further enhance the fidelity and scalability of the proposed framework, multiple extensions can be explored in future research. First, agent movement is currently approximated using Euclidean distances under the assumption of collision-free operation; future work could incorporate explicit collision-avoidance mechanisms alongside optimal path-planning algorithms. Second, orders are currently fulfilled following a FIFO policy. Incorporating order batching and dynamic order-picking strategies could enable agents to select orders based on proximity as well as other operational factors. Finally, battery depletion is assumed to occur at a uniform rate. In practice, energy consumption depends heavily on the carried load; therefore, future directions could consider battery depletion and recharging as functions of the payload.

\section*{Acknowledgments} 
This work was supported by a Research Seed Grant from the Indian Institute of Management Bangalore. The authors also gratefully acknowledge the INFORMS TSL Cross-Regional Collaborative Grant 2025, which initiated this research project.

\bibliographystyle{elsarticle-harv}
\bibliography{references}

\end{document}

%% file: sections/1_introduction.tex
\section{Introduction}

Modern fulfillment centers increasingly rely on Autonomous Mobile Robots (AMRs) to automate intra-logistics operations such as the storage, retrieval, and transportation of inventory. To maintain high throughput in these structured warehouse environments, centralized planning software assigns tasks to these robots, which must be executed efficiently under strict time, energy, and capacity constraints. The scale of these operations is expanding rapidly; for instance, in 2025, Amazon deployed its one-millionth robotic unit \citep{dresser2025amazonrobots}. Their models, such as Titan and Proteus, navigate the warehouse floor, retrieving orders from specific storage locations and transporting them to designated drop-off points, all strictly coordinated by a centralized control system \citep{greenawalt2025amazonrobots}.

The continuous operation of these AMRs is fundamentally constrained by their limited battery capacities. To maintain power, robots must periodically interrupt their assigned tasks and travel to fixed and often inductive or wireless charging stations \citep{wiferionInductiveCharging}.
Commercially used lithium-iron-phosphate batteries require significant downtime. For instance, they average 2.7 hours for a full charge \citep{MCNULTY2022}, while modern models like KUKA's KMP 1500P take approximately one hour to replenish from 20\% to 80\% \citep{KUKA_KMP1500P}. Consequently, charging protocols critically impact warehouse efficiency.
In dense warehouse environments with multiple AMRs and limited charging infrastructure, this creates a complex resource allocation problem. Without coordinated scheduling, multiple robots may seek to charge simultaneously, leading to severe bottlenecks, queuing delays, and an unbalanced utilization of charging stations that ultimately degrades overall system throughput.

The complexity of this charging problem is further amplified in dynamic environments where orders arrive stochastically. In a typical zonal routing setup where individual AMRs are assigned to specific warehouse blocks and must navigate between storage aisles, a central depot, and peripheral charging stations, decision-making is highly interdependent. An AMR must continuously evaluate order selection, depot return timing, and its current state of charge against both immediate charging station availability and anticipated future demand. Traditionally, industry practice has relied on simple rule-based heuristics, such as routing a robot to charge only when its battery falls below a fixed threshold. However, these myopic approaches fail to account for the stochastic nature of order arrivals and the real-time congestion at charging stations, leading to suboptimal fleet coordination.

To address these limitations, we propose a Deep Reinforcement Learning (DRL) framework that empowers AMRs to make dynamic, environment-aware decisions. DRL is well-suited to this task as it effectively manages high-dimensional state spaces and maps complex real-time inputs directly to efficient policies \citep{ref_Mnih2015,Panzer2022}. By capturing the underlying dynamics of the warehouse, our model enables robots to maximize the order fulfillment rate while intelligently balancing battery levels and charging queues. 

The main contributions of this study are as follows: 
\begin{itemize}[noitemsep, topsep=0pt] 
    \item We develop a multi-agent DRL framework utilizing Proximal Policy Optimization (PPO) to jointly optimize routing and charging operations for AMRs in a multi-block warehouse with fixed charging infrastructure. 
    \item We design a comprehensive action space for charging decisions that dictates three critical components: \textit{when} to initiate recharging, \textit{which} charging station to use, and \textit{how long} to spend recharging. 
    \item We incorporate a dynamic drop-off decision model, allowing AMRs to learn when to return picked orders to the depot, thereby minimizing unnecessary travel time. 
    \item We explicitly model the stochastic arrival of orders and the queuing dynamics at charging stations, yielding a coordinated policy that minimizes both overall operational time and unproductive time spent over-charging or waiting. 
\end{itemize}

To validate our approach, we benchmark the proposed DRL model against traditional heuristic methods and alternative DRL baselines, evaluating performance based on order completion rates.

The remainder of this paper is organized as follows. Section 2 reviews the relevant literature, discusses the areas where different DRL techniques are used to improve warehouse operations, and identifies the research gap. In section 3, we describe the problem, the environment, the experimental setup, assumptions, and parameters. In section 4, we formalize the problem statement as a Markov Decision Process (MDP) formulation that will be used to build the DRL framework. In section 5, the methodology and working of the algorithm used for the DRL model is explained. In section 6, we discuss the DRL training results, the policy learned, and evaluate it against other algorithms. In section 7, we perform additional sensitivity analyses to further assess the robustness of the model.

%% file: sections/2_literature_review.tex
\section{Literature Review}
Reinforcement Learning (RL) based approaches have gained increasing attention in warehouse operations, as they enable agents to make real-time decisions and continuously improve performance through interaction with the environment. \citet{KUHNLE2019} discuss a methodical approach to implementing RL algorithms for order dispatching tasks. Recent developments that go beyond traditional manufacturing setups, such as cloud manufacturing, benefit from the deployment of DRL techniques for scheduling tasks \citep{Dong2020}. In this section, we discuss some of the existing literature that uses DRL for warehouse operations. For battery charging operations, we additionally discuss traditional algorithmic approaches alongside DRL frameworks.

\subsection{Agent routing and order picking}
\citet{mahmoudinazlou2022deep} address the dynamic order picking problem, where order arrivals are stochastic and efficient picker routing is crucial, proposing a DRL framework to dynamically optimize picker routes for single-block warehouses. Similarly, \citet{Cals2021} propose a PPO-based order batching and picking framework that reduces the number of late orders, while \citet{beeks2022deep} incorporate an additional objective to minimize picking costs. In a related path-planning study, \citet{Bai2024} develop an improved Double Deep Q-Network (DDQN) model to generate safer and shorter global paths in unknown environments.
To tackle the complexities of dynamic environments, several studies integrate advanced neural architectures with DRL. 
\citet{XiaoGnn2024} integrate a Graph Neural Network (GNN) with PPO to facilitate a multi-agent collaborative navigation system. 
Similarly, using Amazon's Kiva system warehouse layout, \citet{Shen2021} propose a multi-agent Asynchronous Advantage Actor-Critic algorithm combined with an attention mechanism for path planning, while \citet{Li2020a} propose a DQN-based dispatching algorithm that mitigates vehicle traffic while minimizing travel time.
Furthermore, \citet{yang2020impdqn} integrate prior knowledge of collision-free paths with a DQN framework to address slow convergence and excessive randomness in a multi-robot warehouse.
For hierarchical task allocation, \citet{pal2025together} employ a PPO framework to train two agents: a planner that selects tasks from a queue and an executor that assigns the most suitable robot based on its current state, thereby minimizing operation time and execution delays.
 
\subsection{Supply chain inventory management}
DRL has also been applied to inventory management problems, where learning-based approaches optimize storage, replenishment, and production decisions under uncertain demand. To tackle the storage assignment problem, \citet{Teck2025} employ PPO to dynamically reposition storage racks for fast-moving consumer goods, reducing cycle time and congestion at the replenishment station, while \citet{IM07_D3QN_He2023} apply DRL to retrieve requested orders from a dense puzzle-based storage system. Addressing inventory replenishment and demand forecasting problems, \citet{IM18_PPO_Tian2024} combine DRL with Gated Recurrent Unit and multi-head attention to capture demand seasonality and trend from historical observations. Similarly \citet{Chien2020} adopt a DQN model to dynamically select the optimal demand forecast model. Furthermore, in determining order quantity, \citet{Stranieri2024} show that PPO outperforms other DRL algorithms in determining the quantity of products to produce and ship. Although these studies address inventory related optimization problems, they do not consider battery charging decisions for AMRs operating under shared charging constraints.

\subsection{Battery charging}
The optimal scheduling of battery charging for mobile robots has been studied in prior literature, although most existing work only partially aligns with the complexities of multi-block warehouse environments, and many rely on mobile chargers that travel to robots \citep{drenner2009coordinating}. \citet{carli2020control} employ an integer programming framework to learn a battery-charging schedule, while \citet{fu2021online} use mobile chargers that serve robots with low battery, minimizing their meeting time. In a largely stationary environment where robots have less agency, \citet{zou2018evaluating} compare battery recovery strategies: plug-in battery charging, inductive battery charging, and battery swapping, while \citet{tomy2019battery} frame the charging decision as a multi-objective decision-making problem, minimizing battery usage cost and maximizing task completion reward.

\subsubsection{Deep reinforcement learning for battery charging}
Focusing on charging duration, \citet{bischoff2025reinforcement} propose a PPO model that decides the discrete battery level up to which an agent charges at the nearest charging station, while \citet{lin2023real} employs the SARSA algorithm to determine the charging schedule. Studies focusing on agent behavior show that \citet{zellner2022deep} employ a DQN framework to maximize area coverage while minimizing the number of battery constraint violations by agents, and \citet{deng2020battery} develop a framework that selects the number of agents with the lowest battery that need to charge, along with an upper limit at which agents stop charging. In a related study, \citet{zhang2024drl_agv_scheduling} model vehicle scheduling and battery replacement decisions as a bi-objective optimization problem using a Dueling Deep Double Q-network, while \citet{Yoshida2021} learn charging decisions in electric railway trains for improved power utilization.

\subsection{Research Gap}
The reviewed literature highlights the applications of DRL in warehouse optimization, particularly in routing, order picking, and inventory management. However, charging decisions for AMRs remain relatively underexplored in environments where charging resources are shared among multiple AMRs.
Existing approaches generally fail to jointly account for charging station selection, charging duration beyond fixed rules, congestion at charging stations, and dynamic task execution. 
To address these limitations, this study proposes a novel DRL framework designed to maximize order fulfillment rates. Within this framework, AMRs operating in a multi-block warehouse dynamically learn optimal charging strategies while navigating continuous battery depletion, real-time order arrivals, and shared charging infrastructure.

%% file: sections/3_problem_description.tex
\section{Problem Description}

\begin{figure}[!ht]
    \centering
    \includegraphics[width=0.7\linewidth]{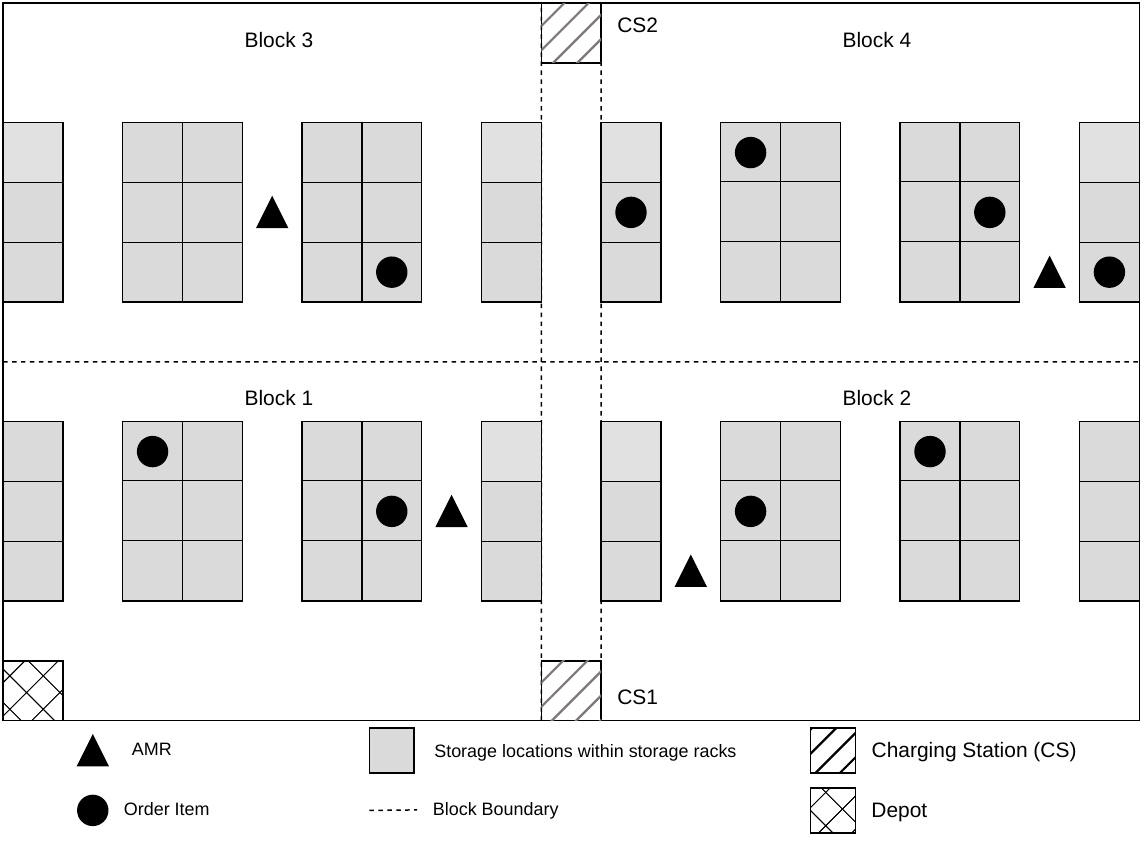}
    \caption{Layout for a warehouse with 4 blocks, 3 aisles per block, 3 storage locations per rack and 2 charging stations}
    \label{fig:warehouse}
\end{figure}

In this section, we formulate a customized warehouse environment to train and evaluate the proposed DRL model. The physical facility is discretized into a coordinate grid to facilitate precise spatial tracking of the AMRs. Hereafter, we refer to an AMR simply as an ``agent''. Consider the warehouse layout illustrated in Figure~\ref{fig:warehouse}.
The warehouse comprises $N$ distinct rectangular blocks, with each block exclusively serviced by a single agent. Customer orders arrive dynamically following a Poisson process with a rate of $\mu$ orders per second. Upon arrival, each order is immediately assigned to the agent responsible for the corresponding block.
The facility includes a depot located at the bottom-left corner for order drop-off, alongside $M$ fixed charging stations distributed evenly along the top and bottom boundaries, where $N > M$. Additionally, each agent has a finite order-carrying capacity, denoted by $K$.

\subsection{Warehouse Environment}

Let the current position of an agent be denoted by the spatial coordinates $(X,Y)$. For the following movements, we define \textit{sufficient charge} as having a battery level equal to the minimum battery threshold $b_\text{min}$ plus the energy required to travel to the next position. While an agent has pending orders to pick up from a storage location, it may perform the following movements based on the current state of the environment:

\begin{enumerate}[noitemsep, topsep=0pt]
    \item Within its block, provided the agent has sufficient charge, it can navigate to a storage location (if it has available carrying capacity), any charging station (CS), or the depot (if it is currently carrying items).
    \item If the agent has reached its maximum carrying capacity, it may travel to the depot, provided it has sufficient charge.
    \item At any point in time, if the agent lacks sufficient charge to reach its intended destination (depot or CS) or if its battery level falls below the predefined minimum threshold $b_\text{min}$, it must travel to a CS.
    \item While at a CS, the agent may begin charging (provided it is at the front of the queue), wait in the queue behind other agents, or stop charging and return to its previous position (provided it has sufficient charge to complete the return trip).
\end{enumerate}

During periods of zero demand, when the order queue is empty, the agent remains idle at its current location to conserve energy. Charging and depot decisions are deferred until new orders arrive.

\subsection{Parameters}
The parameters governing the operational environment, the agents, and the order dynamics are summarized in Table~\ref{tab:cs_parameters}, Table~\ref{tab:agent_parameters}, and Table~\ref{tab:order_parameters}, respectively.
\input{tables/1_s3_parameters}

\subsection{Assumptions}
To balance model complexity with practical relevance, we adopt the following simplifying assumptions:
\begin{itemize}[noitemsep, topsep=0pt]
    \item All blocks, agents and charging stations are homogeneous sharing identical physical specifications, differing only in their spatial locations, dynamic agent behaviors, and real-time queue lengths.
    \item Battery depletion occurs only during transit. Energy consumption during idle periods (e.g., waiting in a charging queue, charging, or remaining stationary due to a lack of orders) is considered negligible.
    \item The minimum battery threshold $b_\text{min}$ is sufficient to ensure that any agent can reach the nearest CS from any location within its operational range, thereby preventing complete battery depletion during operations.
    \item Constant battery depletion rate $\eta$ is assumed for simplicity and to facilitate benchmarking with existing studies.
    \item Agents navigating to common destinations are assumed to be collision-free.
\end{itemize}

%% file: tables/1_s3_parameters.tex
\begin{table}[!ht] \centering
\caption{Charging station related parameters.}
\label{tab:cs_parameters}
\begin{tabular}{ll}
\toprule
\textbf{Parameter} & \textbf{Description} \\
\midrule
$M$ & Number of charging stations (CS)\\
$\mathcal{M}$      & Set of $M$ CS, individually denoted as CS$m$ (e.g.,$CS1$, $CS2$ for $M=2$)\\
$Q_t^m$            & Set of agents waiting in queue at the $m^{\text{th}}$ CS at time $t$\\
$q_t^m = |Q_t^m|$  & Queue length of $m^{\text{th}}$ CS at time $t$\\
\bottomrule
\end{tabular}
\end{table}

\begin{table}[!ht]
\centering
\caption{Agent related parameters.}
\label{tab:agent_parameters}
\begin{tabularx}{\textwidth}{lX}
\toprule
\textbf{Parameter} & \textbf{Description} \\
\midrule
\multicolumn{2}{l}{\textit{Static parameters}} \\
\midrule
$N$                            & Number of agents with their individual blocks \\
$N_{\text{aisle}}$             & Number of aisles per block \\
$N_{\text{slot}}$              & Number of storage locations on either side of an aisle\\
$\mathcal{N}$                  & Set of $N$ agents, individually denoted as $An$ (e.g.,$A1$, $A2$, $A3$, $A4$ for $N=4$)\\
$K$                            & Maximum carrying capacity of an agent \\
$b_\text{max}$                 & Maximum battery level (unit) \\
$b_\text{min}$                 & Minimum battery threshold (unit) \\
$\eta$                         & Battery depletion rate (unit/sec) \\
$\beta$                        & Battery re-charge rate (unit/sec) \\
$V$                            & Velocity of the agent (unit/sec) \\
$d_{\text{max}}$               & ($N_{\text{aisle}}$ + $N_{\text{slot}}$) $\times$ 2; Hyperparameter used for normalizing distances\\
\midrule
\multicolumn{2}{l}{\textit{Dynamic parameters at time step $t$}} \\
\midrule
$(x_t^{n},y_t^{n})$            & Current location (aisle, storage location) of agent $n$\\
$d_t^{n,o}$                    & Normalized Euclidean distance from agent $n$ to its next order; $-1$ if no next order available \\
$d_t^{n,1}, \dots, d_t^{n,M}$ & Normalized Euclidean distances from agent $n$ to $M$ charging stations \\
$d_t^{n,d}$                    & Normalized Euclidean distance from agent $n$ to the depot \\
$d_t^{i,j}$                    & Normalized Euclidean distance from agent $i$ to agent $j$ \\
$k_t^n$                        & Current carrying capacity of agent $n$ \\
$b_t^n$                        & Current battery charge of agent $n$ (unit) \\
$\sigma_t^n$                   & Service mode of agent $n$; equals $m$ if at $m^{\text{th}}$ CS, else $0$ \\
\bottomrule
\end{tabularx}
\end{table}

\begin{table}[!ht] \centering
\caption{Order related parameters.}
\label{tab:order_parameters}
\begin{tabular}{ll}
\toprule
\textbf{Parameter} & \textbf{Description} \\
\midrule
$o^i = (n^i, x^i, y^i, k^i, p^i, i)$ & Representation of the requested order $i$\\
$n^i$ & Block index associated with the requested order $i$ \\
$x^i,y^i$ & Aisle and storage location position of the requested order $i$\\
$\kappa^i$ & Number of items requested from position $x^i,y^i$\\
$p^i \in \{0,1\}$ & Binary indicator: 1 if order $i$ is picked, otherwise $0$\\
$i$ & Order index \\
$\mu$ & Order arrival rate that follows a poisson process \\
\bottomrule
\end{tabular}
\end{table}

%% file: sections/4_mdp_formulation.tex
\section{Markov Decision Process Formulation}

To capture the sequential and stochastic nature of dynamic charging for order pickers, we formulate the problem as a Markov Decision Process (MDP). An MDP provides a rigorous mathematical framework for modeling decision-making under uncertainty. Formally, the MDP is defined by a tuple $(S, A, P, R, \gamma)$, where $S$ represents the set of environment states and $A$ denotes the action space available to the agent. The state transition probability $P$ denotes the probability of moving to the next state given the current state and action, $R$ is the immediate reward function indicating the reward received for taking an action in the current state, and $\gamma$ is the discount factor that takes a value between 0 and 1. A higher value of $\gamma$ corresponds to prioritizing long-term goals.

\textbf{State:} The local observation ${s}_t^n$ of agent $n$ at time step $t$ consists of three components: the agent's own state, the CS queue state, and the states of all other agents. Knowing the state of the neighboring agents aids in better coordination. The agent $n$'s own state includes its distances to its next order to pick, all charging stations, and depot ($d_t^{n,o},\; d_t^{n,1}, \ldots, d_t^{n,M},\; d_t^{n,d}$), normalized battery level $b_t^n/b_{\text{max}}$, capacity level $k_t^n/K$ and service mode  $\sigma_t^n/M$. For the remaining agents, we include their normalized battery levels, capacities, and service modes, alongside their normalized Euclidean distance $d_t^{i,j}$ from agent $n$. In addition, the state includes the queue length $q_t^m$ at charging station $m$, for all charging stations.

\textbf{Action:} At each decision point, an agent $n$ selects an action 
$a_t^n$ from a finite action set $\mathcal{A}$, where $|\mathcal{A}| = M + 6$. 
Table~\ref{tab:actions} provides detailed descriptions of the various actions available to the agents.

\input{tables/2_s4_action}

To ensure operational feasibility and prevent agents from executing illegal actions, the choice is constrained by a binary action mask $\mathbf{m}_t^n \in \{0,1\}^{|\mathcal{A}|}$.
This mask is dynamically computed by the environment at each decision step and enforces the following rule set:
\begin{itemize}[noitemsep, topsep=0pt]
    \item If an agent is en route to a location, only the \texttt{travelling} action is permitted.
    \item If the agent has sufficient capacity and battery energy to service the next order, it can choose among \texttt{go\_pick}, \texttt{go\_to\_CS}, or \texttt{go\_to\_depot}.
    \item If the battery level falls below a minimum threshold, or is insufficient to reach the next task location, only \texttt{go\_to\_CS} actions are allowed.
    \item When the agent is first in the queue at a CS and $b_t^n < b_{\text{max}}$, it can either \texttt{stop\_charging} or \texttt{keep\_charging}.
    \item When $b_t^n = b_{\text{max}}$, only the \texttt{stop\_charging} action is permitted.
    \item When the agent is at a charging station but not first in the queue, only the \texttt{wait\_in\_queue} action is permitted.
\end{itemize}
These hard constraints substantially reduce the effective action space at most decision points, allowing the policy to focus on non-trivial decisions: when to initiate charging, which station to select, when to terminate charging, and when to pick orders or return to the depot.

\textbf{Reward:} The per-agent reward at each time step is designed to maximize order throughput while minimizing non-productive time. Formally, the reward $R_t^n$ for agent $n$ at time step $t$ is defined as follows:

\begin{itemize}[noitemsep, topsep=0pt]
  \item Pick initiation reward $R_\text{pick}$:
    An intermediate reward of $+20$ is allocated when an agent initiates a pick action. Because the primary objective of order completion is observed only sparsely, this reward-shaping term provides an immediate feedback signal to guide the agent toward productive order-picking behaviors.
  \item Per-step penalty $R_\text{step}$:
    A penalty of  $-1$ is applied at every time step, regardless of the action taken. This encourages the agent to minimize the overall time required to fulfill orders, including unnecessary waiting, prolonged charging and inefficient travel.
\end{itemize}

We do not enforce explicit charging-threshold rewards or depot-utilization bonuses, expecting the agent to discover the optimal charging timing, charging duration, and depot batching policy entirely from the interplay between the per-step opportunity cost and the order-completion signal. Our preliminary experiments explored more complex reward formulations by accounting for time spent charging, operational time lost, completed orders, items delivered, and queue lengths to explicitly guide these trade-offs, minimize delays, and penalize premature visits. However, numerical results demonstrated that the simpler reward structure achieved superior sample efficiency and policy performance.

\textbf{State Transitions:} State transitions are evaluated at every global time step $t$.
Each agent $n \in \mathcal{N}$ maintains a countdown timer tracking the duration of its current travel segment. During this period, the agent is locked out of new decisions; once it completes this segment, it is free to select its next action.

\paragraph{Agent position}
During travel, the position of agent $n$ is updated by linear interpolation:
\begin{equation}
  x_t^n = x_{\text{orig}} + \tau_t \left(x_{\text{dest}} - x_{\text{orig}}\right), \quad
  y_t^n = y_{\text{orig}} + \tau_t \left(y_{\text{dest}} - y_{\text{orig}}\right),
\end{equation}
where $(x_{\text{orig}}, y_{\text{orig}})$ and $(x_{\text{dest}}, y_{\text{dest}})$ denote the spatial coordinates of the agent's origin and destination, respectively. $\tau_t = (T_{\text{total}} - T_{\text{rem}}) / T_{\text{total}}$ represents the portion of the journey completed, with $T_{\text{total}}$ being the total travel time to the next location and $T_{\text{rem}}$ the remaining travel time. Depot trips follow a two-phase interpolation: the position moves toward the depot for the first half of the allotted time and returns from the depot during the second half. Hence, a depot trip begins when an agent decides to travel to the depot and ends after all the items carried have been unloaded and the agent has returned to its departure location. Only then is the agent allowed to make the next decision.

\paragraph{Battery}
The battery depletes strictly during travel, proportional to distance traveled as follows:
\begin{equation}
  b_{t+1}^n = b_t^n - \eta \cdot \Delta d_t^n,
\end{equation}

where $\eta$ is the depletion rate and $\Delta d_t^n$ is the
Euclidean distance traversed in the current step; however, when the agent is first in queue and executes \texttt{keep\_charging}, the battery updates as:
\begin{equation}
  b_{t+1}^n = \min\!\left(b_{\text{max}},\; b_t^n + \beta\right),
\end{equation}
where $\beta$ is the recharge rate per step.
The battery does not deplete while idle, queuing, or charging.

\paragraph{Capacity and order completion}
Upon arrival at an order location, the order is marked as complete, the agent's carrying capacity decreases by the order's demand $\kappa_i$, and the order is removed from the active list. Upon completion of a depot round trip, the agent's carrying capacity is fully restored to its maximum, $K$.

\paragraph{CS queue}
When an agent arrives at charging station $m$, it joins the station's queue $Q^m$, and the total queue length $q^m$ increments by one. Only the agent at the front of the queue is permitted to charge. Upon executing the \texttt{stop\_charging} action, the agent is dequeued, and $q^m$ decrements by one.

\paragraph{Order arrivals}
Orders arrive according to a Poisson process with rate $\mu$, and each new order is uniformly assigned to one of the $N$ operational blocks.

%% file: tables/2_s4_action.tex
\begin{table}[!ht] \centering
\caption{Action space $\mathcal{A}$ for each agent.}
\label{tab:actions}
\begin{tabularx}{\textwidth}{lX}
\hline
\textbf{Action Name} & \textbf{Description} \\
\hline
\texttt{go\_pick} & Go to the next order and pick it.\\
\texttt{go\_to\_CS} & Go to charging station $m$, $\forall m \in \mathcal{M}$. \\
\texttt{stop\_charging} & Stop charging and return; available only when the agent is first in queue at CS. \\
\texttt{go\_to\_depot} & Go to depot to replenish capacity and mandatorily return to the departure position; available when $k_t^n < K$.\\
\texttt{wait\_in\_queue} & Enforced by the environment while the agent is not first in queue at a CS. \\
\texttt{keep\_charging} & Enforced by the environment when agent is at CS and \texttt{stop\_charging} action is not taken.\\
\texttt{travelling} & Enforced by the environment during travel. \\
\hline
\end{tabularx}
\end{table}

%% file: sections/5_methodology_PPO.tex
\section{Methodology}

We formulate the warehouse order fulfillment process as a Multi-Agent Reinforcement Learning (MARL) problem and solve it using Independent Proximal Policy Optimization (IPPO) with a shared actor and per-agent independent critics. At each time step $t$, agent $n$ observes a local state $s_t^n$, selects an action $a_t^n$, and receives a scalar reward $r_t^n$.

In IPPO, each agent learns its own policy and value function using only local observations. We follow the standard MARL convention, in which ``local observation'' denotes that the observation is agent-specific and computed without global coordination, not that it is restricted to own-agent features. Including neighboring agents' battery levels, capacities, and relative distances in local observations is analogous to an agent perceiving its environment through onboard sensors: the information is available locally at execution time without any centralized aggregation. Unlike centralized-training approaches (e.g., MAPPO with a joint critic), IPPO maintains full decentralization: no agent conditions its critic on the joint observation $[s_t^1, \ldots, s_t^N]$.

All agents share a single actor network that maps local observations to action probabilities, while each agent maintains an independent critic network that inputs the agent's local observations and estimates the value of its current state. Value of a state gives information regarding how much cumulative reward an agent can be expected to collect if the agent starts from the current state and progresses forward in time under the current policy. During each episode, agents interact with the environment, collect states, actions, and rewards. At the end of each episode, Temporal-Difference (TD) errors are computed for every state visited. The TD error at each state measures whether the reward received at that state plus the critic's estimate of the next state (the state resulting from the action taken) was higher or lower than the critic's estimate of the current state. A positive TD error indicates the outcome was better than the critic's expectation; a negative TD error indicates otherwise. For all states, TD errors of the current and subsequent time steps are accumulated using the Generalized Advantage Estimation (GAE) framework to produce advantages, which reflect the long-term consequences of each action and inform whether the action taken at each state led to a better-than-expected outcome. These advantage values, when added to the critic's current estimate, give a better estimate of the returns that can be obtained from a state. The returns thus obtained are used to calculate the critic loss. For the actor network, PPO computes the ratio of the new policy's probability of taking an action to the old policy's probability; this ratio is clipped to prevent drastic changes, and multiplied by the advantage, thereby providing the magnitude and direction of the policy update. The following subsections explain each component of the IPPO model in detail.

\subsection{Average Reward Formulation}

\begin{figure}[!ht]
    \centering
    \begin{subfigure}{0.49\textwidth}
        \centering
        \includegraphics[width=\linewidth]{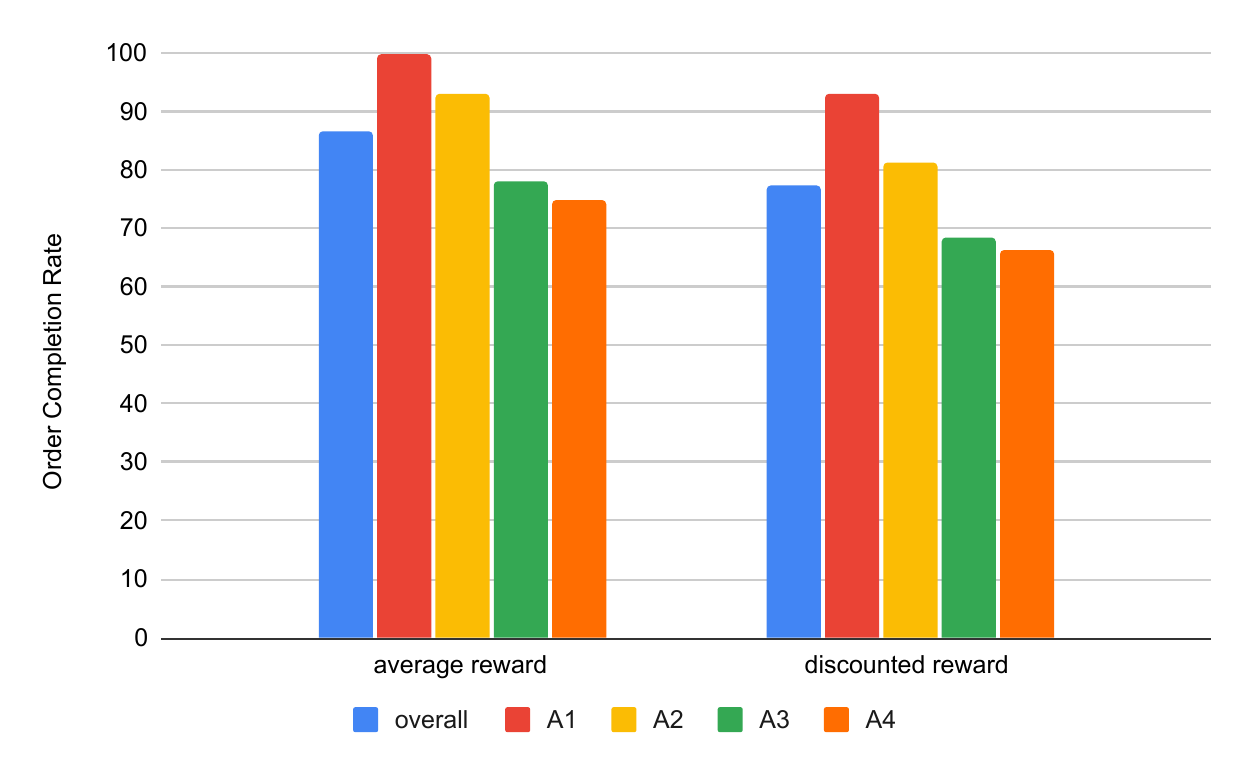}
    \end{subfigure}
    \caption{Order completion rate when using average reward versus discounted reward in a representative environment}
    \label{fig:s5_1}
\end{figure}

A key design choice in our framework is the use of an average reward formulation in place of the standard discounted return.
Standard PPO relies on a discount factor $\gamma < 1$, which geometrically downweights future rewards:
\begin{equation}
J_\gamma = \mathbb{E}_\pi\left[\sum_{t=0}^{\infty} \gamma^t r_t\right].
\end{equation}
In long-horizon or continuing tasks, this causes the agent to heavily discount rewards that occur far later in the episode. However, in our warehouse environment, every order arrival is equally important to fulfill. \citet{av_rew2022} demonstrated that in a warehouse AMR battery management task, discounted-reward algorithms plateaued below a 90\% order completion rate, while their average-reward counterpart consistently exceeded 95\%. We observe similar trends for our problem, as shown in Figure~\ref{fig:s5_1}.

Consequently, we maximize the long-run average reward per timestep under a policy $\pi$:
\begin{equation}
\rho_\pi^n = \lim_{T \to \infty} \frac{1}{T} \mathbb{E}_\pi \left[ \sum_{t=0}^{T-1} r_t^n \right],
\label{eq:rho_definition}
\end{equation}
where $r_t^n$ is the reward received by agent $n$ at timestep $t$ under policy $\pi$. By adopting this formulation, we remove $\gamma$ as a hyperparameter and ensure a uniform weighting of all time steps.

Without a discount factor, cumulative returns in continuing tasks grow to infinity. To stabilize training, we maintain a baseline $\bar{r}^n$, which serves as a running estimate of the long-term average reward $\rho_\pi^n$ for each agent $n$. This estimate is updated at the end of each episode $e$ via an exponential moving average:
\begin{equation}
\bar{r}_{e}^n = (1 - \alpha_{\bar{r}}) \bar{r}_{e-1}^n + \alpha_{\bar{r}} \cdot \frac{1}{T} \sum_{t=0}^{T-1} r_t^n,
\label{eq:rbar_update}
\end{equation}
where $T$ is the episode duration, $\alpha_{\bar{r}}$ is the update rate, and the sum represents the total reward accumulated by agent $n$ during episode $e$. Initialized at $\bar{r}_0^n = 0$, this formulation provides a smooth, adaptive baseline. 
After updating, this baseline $\bar{r}^n$ is subtracted from the immediate rewards to compute the TD error in Eq.~\eqref{eq:td_error}.

\subsection{Network Architecture}

\textbf{Shared Actor Network:} All agents share a single actor network parameterized by $\theta$. To allow agent-specific policies, the local observation $s_t^n$ is concatenated with a one-hot encoding of the agent identity $u^n \in \{0,1\}^N$ before being passed through the network. The actor maps $(s_t^n, u^n)$ to action logits, from which the action distribution $\pi_\theta(a \mid s_t^n, u^n)$ is derived by softmax. At each decision point, infeasible actions (determined by action mask $\mathbf{m}_t^n$) are masked by setting their logits to $-10^9$ before the softmax. During training, actions are sampled from this masked distribution; during evaluation, the argmax of the masked logits is used.

\input{tables/3_s5_actornetwork}

\textbf{Independent Local Critics:} Each agent $n$ maintains a separate critic network $V_{\phi^n}$ with parameters $\phi^n$ that are trained independently. The network approximates the \textit{differential} state value function from the agent's local observation:
\begin{equation}
V_{\phi^n}(s_t^n) \approx V_\pi^n(s_t^n) := \mathbb{E}_\pi\left[\sum_{\tau=t}^{\infty} (r_\tau^n - \rho_\pi^n) \,\Big|\, s_t^n\right],
\label{eq:value_function}
\end{equation}
which measures the expected cumulative deviation from the agent's long-run average reward $\rho_\pi^n$. This defines the \textit{differential} value function, which reflects the expected relative performance of state $s_t^n$ compared to the stationary average reward baseline. Each critic takes the local observation as input and outputs a scalar value.

\input{tables/4_s5_criticnetwork}

\subsection{Advantage Estimation via Differential GAE}

Under the average-reward formulation, the TD error at timestep $t$ for agent $n$ is defined as:
\begin{equation}
\delta_t^n = r_t^n - \bar{r}^n + V_{\phi^n}(s_{t+1}^n) - V_{\phi^n}(s_t^n).
\label{eq:td_error}
\end{equation}
Eq.~\eqref{eq:td_error} captures the one-step prediction error relative to the running average baseline $\bar{r}^n$.  This is the average-reward analog of the standard discounted TD error $r_t + \gamma V(s_{t+1}) - V(s_t)$ with $\gamma=1$.

To reduce variance, we compute advantages using Differential GAE, which accumulates TD errors backward in time with exponential decay parameter $\lambda$:
\begin{equation}
\hat{A}_t^n = \sum_{\ell=0}^{T-t-1} \lambda^\ell \delta_{t+\ell}^n.
\label{eq:gae}
\end{equation}
This exponentially weights recent TD errors more heavily, with $\lambda$ providing a balance between bias and variance. The corresponding return target for critic training is:
\begin{equation}
G_t^n = \hat{A}_t^n + V_{\phi^n}(s_t^n).
\label{eq:returns}
\end{equation}
Finally, to stabilize the actor network updates, the estimated advantages are normalized per agent to maintain a zero mean and unit variance across the episode.

\subsection{PPO Objectives}
\textbf{Actor Loss:} The actor is updated using the PPO clipped surrogate objective. Let $\text{ratio}_t^n(\theta)$ denote the importance sampling ratio:
\begin{equation}
\text{ratio}_t^n(\theta) = \frac{\pi_\theta(a_t^n \mid s_t^n, u^n)}{\pi_{\theta_{\text{old}}}(a_t^n \mid s_t^n, u^n)},
\end{equation}
where $\pi_\theta(a_t^n \mid s_t^n, u^n)$ denotes the probability assigned by the current policy to action $a_t^n$ given the agent's local observation $s_t^n$ and one-hot encoding of the agent identity $u^n \in \{0,1\}^N$. The observation and identity encoding are concatenated and passed as input to the the actor network. Here, $\theta_{\text{old}}$ are the actor parameters from the previous policy. The sampling ratio measures how much the current policy has changed relative to the old policy for the specific action taken. The actor loss which is a clipped objective with entropy regularization is:
\begin{equation}
L^{\text{actor}}(\theta) = -\mathbb{E}\left[\min\left(\text{ratio}_t^n \cdot \hat{A}_t^n,\; \text{clip}(\text{ratio}_t^n, 1-\varepsilon, 1+\varepsilon) \hat{A}_t^n\right)\right] - \alpha_H \cdot H[\pi_\theta],
\label{eq:actor_loss}
\end{equation}
The loss is negated because the goal is to maximize the expected return whereas optimizers (Adam Optimizer) perform gradient descent which minimizes it. The first term ensures that when the action taken was better than expected, leading to a positive advantage ${A}_t^n$, an increase in the probability of that action $\text{ratio}_t^n$ is encouraged, and when the advantage is negative it is discouraged. However, if the  sampling ratio is allowed to change without a bound, a single update could change the policy drastically causing instability in the policy. To prevent this, the ratio is restricted to stay within the clipping threshold of $[1-\varepsilon, 1+\varepsilon]$. Taking a minimum of the clipped and unclipped objectives prevents potentially harmful updates. In the second term, the policy entropy $H[\pi_\theta] = -\sum_a \pi_\theta(a \mid s_t^n, u^n) \log \pi_\theta(a \mid s_t^n, u^n)$  measures the randomness of the action distribution. A higher value means actions are assigned a more uniform probabilities and a lower entropy means actions are more deterministic. Since the loss subtracts $\alpha_H \cdot H[\pi_\theta]$, minimizing the loss is equivalent to maximizing the entropy, encouraging diverse action probabilities and thus preventing early convergence. To balance exploration early in training and exploitation later, the entropy coefficient $\alpha_H$ decays linearly as:
\begin{equation}
\alpha_H = \max\left(0.01,\; H_0 \cdot \left(1 - \frac{e}{7000}\right)\right),
\end{equation}
where $e$ is the episode number and ${H_0}$ is the starting entropy coefficient. This encourages exploration early in training and exploitation later. The value of 7000 was chosen empirically to maintain exploration during the majority of training, approximately 70\% of the 10,000 training episodes, before gradually shifting toward exploitation during the final training episodes. The actor is updated for $N_{\text{actor\_update}}$ epochs per episode using randomly shuffled mini-batches of size 512, drawing samples from all agents jointly. All agent samples within a mini-batch are processed in a single forward pass through the shared actor.\par

\textbf{Critic Loss:} Each critic is trained independently to minimize the Huber loss between predicted values $V_{\phi^n}(s_t^n)$ and return targets $G_t^n$ computed from 
Eq.~\eqref{eq:returns}:
\begin{equation}
L^{\text{critic}}(\phi^n) = \mathbb{E}\left[L_\delta\left(V_{\phi^n}(s_t^n),\; G_t^n\right)\right],
\label{eq:critic_loss}
\end{equation}
where $L_\delta$ is the Huber loss. Unlike Mean Squared Error (MSE) which squares large errors that occur during initial training, Huber loss transitions from quadratic to linear behavior beyond a prediction error threshold $\delta$:
\begin{equation}
L_\delta(y, \hat{y}) = \begin{cases}
\frac{1}{2}(y - \hat{y})^2 & \text{if } |y - \hat{y}| \leq \delta, \\
\delta \cdot (|y - \hat{y}| - \frac{1}{2}\delta) & \text{otherwise}.
\end{cases}
\end{equation}
where $y = G_t^n$ is the return target, $\hat{y} = V_{\phi^n}(s_t^n)$ is the critic's prediction, and $\delta = 1.0$ is the threshold that determines where the transition from quadratic to linear occurs. Each critic is updated for $N_{\text{critic\_update}}$ epochs per episode. For each mini-batch, agent samples are extracted via an agent-index mask, and each critic is updated independently on its own samples only.

\subsection{Training Procedure}

Algorithm~\ref{alg:ippo} outlines the complete training process, where each episode consists of five distinct phases:
\begin{enumerate}[noitemsep, topsep=0pt]
    \item \textbf{Rollout collection:} At each timestep $t$, each agent $n$ computes $s_t^n$ from the environment observation, computes its action mask $\mathbf{m}_t^n$, and samples an action $a_t^n \sim \pi_\theta(\cdot \mid s_t^n, u^n) \odot \mathbf{m}_t^n$.
    The tuple $(s_t^n, a_t^n, \log\pi_\theta(a_t^n), r_t^n, \mathbf{m}_t^n)$ is appended to a rollout buffer $\mathcal{D}$ of capacity $T \times N$ (episode length times number of agents).
    Inter-agent communications do not occur during the rollout.

    \item \textbf{Advantage computation:} Upon episode completion, each agent's critic evaluates its collected trajectory to produce value estimates $\{V_{\phi^n}(s_t^n)\}_{t=0}^{T-1}$. We then apply Differential GAE (Eq.~\ref{eq:gae}), followed by per-agent advantage normalization. Finally, the return targets $G_t^n$ are computed for the upcoming critic training (Eq.~\ref{eq:returns}).

    \item \textbf{Critic updates:} The buffer $\mathcal{D}$ is flattened into $T \times N$ samples. Over $N_{\text{critic\_update}}$ epochs, these samples are shuffled and divided into mini-batches of size 512. Within each mini-batch, each independent critic is updated using only the subset of samples belonging to its agent. Gradient norms are clipped to 0.5 before each parameter update.

    \item \textbf{Actor updates:} The same flattened and shuffled buffer is reused for the policy updates. Over $N_{\text{actor\_update}}$ epochs, mini-batches of size 512 are drawn. All samples within a mini-batch are passed through the shared actor network in a single forward pass, yielding updated log-probabilities and entropies. As with the critic, gradient norms are clipped to 0.5 before parameter updates.

    \item \textbf{Baseline updates:} Following the network parameter updates, the per-agent average reward baselines $\bar{r}^n$ are updated via Eq.~\ref{eq:rbar_update}, using the mean reward of the current episode.
\end{enumerate}

\begin{algorithm*}[!ht]
\caption{IPPO Training Loop}\label{alg:ippo}
\begin{algorithmic}[1]
\State \textbf{Initialize:} Actor $\pi_\theta$, critics $\{V_{\phi^n}\}_{n=1}^N$,
       baselines $\{\bar{r}^n=0\}_{n=1}^N$
\For{episode $e = 1, 2, \ldots, E$}
    \Statex \textbf{Phase 1: Rollout Collection}
    \State Reset environment; initialise buffer $\mathcal{D} \gets \emptyset$
    \For{$t = 0$ to $T-1$}
        \For{agent $n = 1$ to $N$}
            \State Observe $s_t^n$; compute action mask $\mathbf{m}_t^n$
            \State Sample $a_t^n \sim \pi_\theta(\cdot \mid s_t^n, u^n) \odot \mathbf{m}_t^n$;
                   store $\log\pi_\theta(a_t^n)$
        \EndFor
        \State Execute joint action; observe rewards $\{r_t^n\}$; append to $\mathcal{D}$
    \EndFor
    \Statex \textbf{Phase 2: Advantage Computation}
    \For{agent $n = 1$ to $N$}
        \State Forward pass: $\{V_{\phi^n}(s_t^n)\}_{t=0}^{T-1}$
        \State Compute $\{\hat{A}_t^n\}$ via Eqs.~\eqref{eq:td_error}--\eqref{eq:gae};
               normalise per-agent
        \State Compute returns: $G_t^n = \hat{A}_t^n + V_{\phi^n}(s_t^n)$
    \EndFor
    \Statex \textbf{Phase 3: Critic Update ($N_{\text{critic}}$ epochs)}
    \For{epoch $1$ to $N_{\text{critic}}$}
        \State Shuffle $\mathcal{D}$; partition into mini-batches
        \For{each mini-batch}
            \For{agent $n = 1$ to $N$}
                \State Extract agent-$n$ samples; compute Eq.~\eqref{eq:critic_loss};
                       clip gradients; update $\phi^n$
            \EndFor
        \EndFor
    \EndFor
    \Statex \textbf{Phase 4: Actor Update ($N_{\text{actor}}$ epochs)}
    \For{epoch $1$ to $N_{\text{actor}}$}
        \State Shuffle $\mathcal{D}$; partition into mini-batches
        \For{each mini-batch}
            \State Forward pass all samples through shared actor
            \State Compute Eq.~\eqref{eq:actor_loss} on samples;
                   clip gradients; update $\theta$
        \EndFor
    \EndFor
    \Statex \textbf{Phase 5: Baseline Update}
    \For{agent $n = 1$ to $N$}
        \State $\bar{r}_k^n \gets (1-\alpha_{\bar{r}})\bar{r}_{k-1}^n
               + \alpha_{\bar{r}} \cdot \tfrac{1}{T}\sum_{t}r_t^n$
    \EndFor
\EndFor
\end{algorithmic}
\end{algorithm*}

We adopt the following key design choices. The critic network undergoes more update epochs than the actor to ensure robust value estimates for advantage computation while preventing the actor from overfitting to stale rollout trajectories. Furthermore, layer normalization is applied after each hidden layer in both networks to stabilize training across heterogeneous state inputs (e.g., normalized distances, battery ratios, and integer queue lengths) that span different scales. To avoid distortions in gradient signal, normalizing advantages independently for every agent preserves individual reward scales preventing agents experiencing high-variance episodes from dominating the global gradient signal. Processing samples simultaneously from all agents in a single forward pass per mini-batch also ensures that balanced gradient signals from all agents are received by the shared actor network. In addition, gradients of both networks are clipped to a maximum $\ell_2$ norm of 0.5. This mitigates training instability caused by excessively large policy updates when immediate rewards fluctuate significantly.

%% file: tables/3_s5_actornetwork.tex
\begin{table}[!ht] 
\centering
\caption{Shared actor network architecture. All $N$ agents use the same network weights.}
\label{tab:actor}
\begin{tabular}{ll}
\toprule
\textbf{Layer} & \textbf{Specification} \\
\midrule
Input    & Local obs. $\oplus$ agent one-hot \\
Hidden 1 & FC: Input $\to 512$, ReLU, LayerNorm \\
Hidden 2 & FC: $512 \to 512$, ReLU, LayerNorm \\
Hidden 3 & FC: $512 \to 256$, ReLU \\
Output   & FC: $256 \to$ Action logits \\
\bottomrule
\end{tabular}
\end{table}

%% file: tables/4_s5_criticnetwork.tex
\begin{table}[!ht] 
\centering
\caption{Local critic network architecture. Each of the $N$ agents maintains a separate independent critic.}
\label{tab:critic}
\begin{tabular}{ll}
\hline
\textbf{Layer} & \textbf{Specification} \\
\hline
Input         & Local obs. \\
Hidden 1      & FC: Input $\to 256$, ReLU, LayerNorm \\
Hidden 2      & FC: $256 \to 256$, ReLU, LayerNorm \\
Hidden 3      & FC: $256 \to 128$, ReLU \\
Output        & FC: $128 \to 1$ scalar \\
\hline
\end{tabular}
\end{table}

%% file: sections/6_results_PPO.tex
\section{Experiments}
We conduct experiments using two different setups, where the primary difference is their spatial scale: Environment E-I represents a 4-block warehouse, whereas E-II represents a 6-block warehouse. Training was conducted over 10,000 episodes during a 4-hour simulation, while testing was conducted in an 8-hour simulation. The source code is available on GitHub: \url{https://github.com/taniya-0/dynamic-battery-drl/}.

\begin{figure*}[!ht]
    \centering
    \begin{subfigure}{0.85\textwidth}
        \centering
        \includegraphics[width=\linewidth]{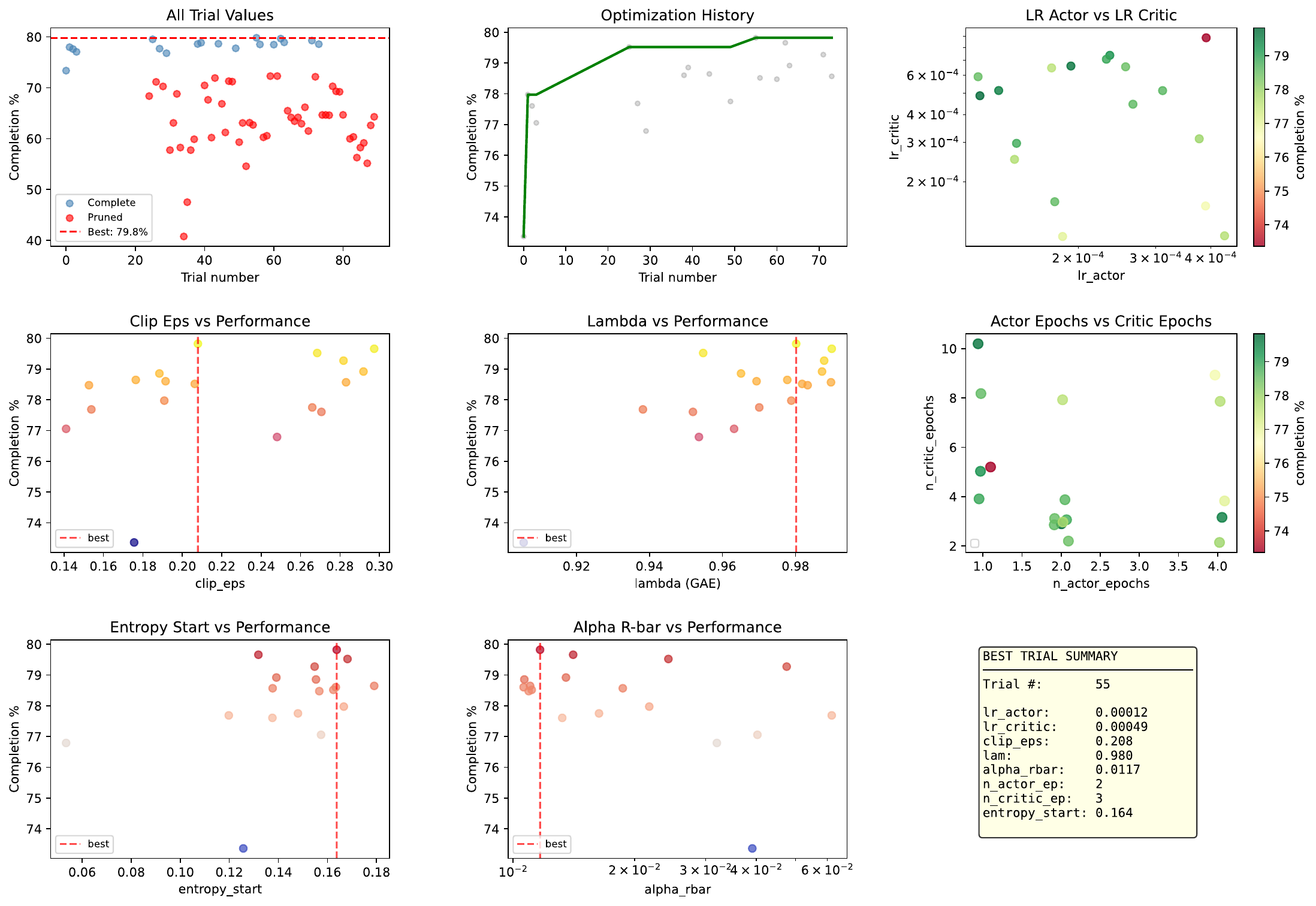}
    \end{subfigure}
    \caption{Hyperparameter tuning results performed in E-I environment}
    \label{fig:s4_1}
\end{figure*}

Hyperparameter tuning was performed using Bayesian optimization via the Optuna framework \citep{akiba2019optuna}, which leverages the Tree-structured Parzen Estimator (TPE) sampler. Rather than exhaustively searching for all combinations of the parameter, TPE iteratively focuses sampling on promising regions of the search space based on the results of previous outcomes. Table~\ref{tab:experiment_settings} reports the environment parameters, hyperparameter search range and final tuned values used in our experiments. Figure~\ref{fig:s4_1} summarizes the hyperparameter tuning results in E-I environment.

\input{tables/5_s6_env_hyper_params}

\subsection{Training Results}
\begin{figure*}[!ht]
    \centering
    \begin{subfigure}{0.45\textwidth}
        \centering
        \includegraphics[width=\linewidth]{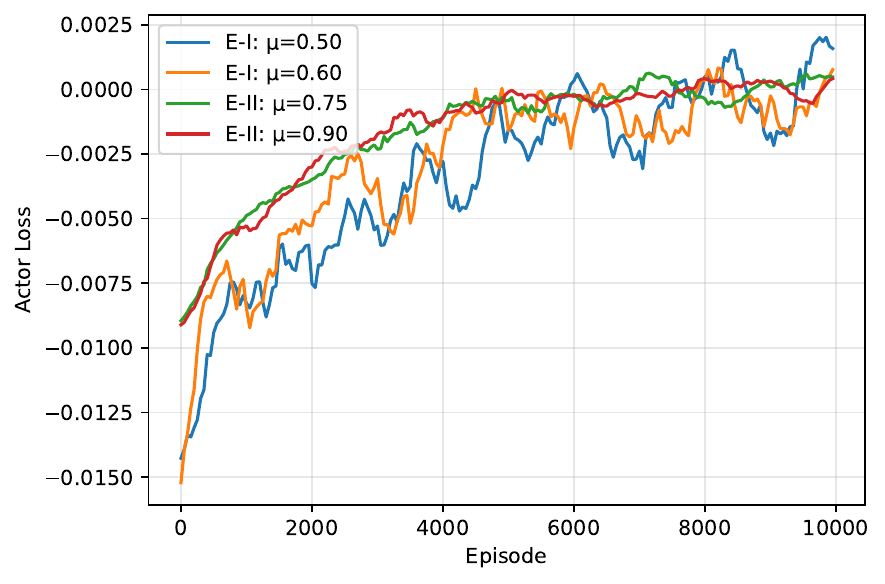}
        \caption{Actor Loss}
    \end{subfigure}
    \hfill
    \begin{subfigure}{0.45\textwidth}
        \centering
        \includegraphics[width=\linewidth]{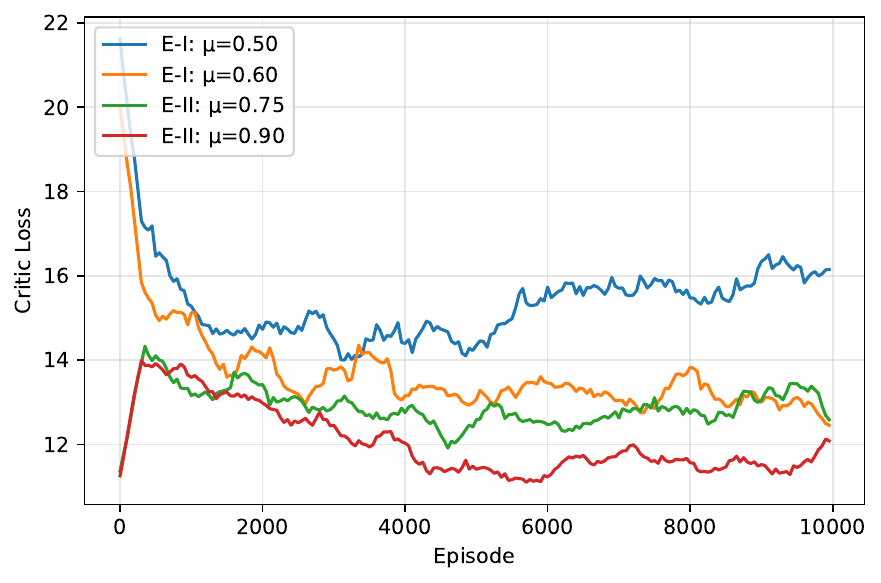}
        \caption{Critic Loss}
    \end{subfigure}
    \caption{Propagation of actor and critic losses during training under different scenarios.}
    \label{fig:s6_1}
\end{figure*}

Figure~\ref{fig:s6_1} illustrates the actor and critic losses over the course of training. The actor loss is defined as the negative of the PPO clipped surrogate objective given in Eq.~\eqref{eq:actor_loss}. The first term of this objective quantifies the degree of policy improvement, while the second term measures the policy's exploration. At the onset of training, substantial policy improvements and high exploration rates drive the actor loss to be strongly negative. As the policy converges, the importance sampling ratio approaches $1.0$ and the entropy coefficient decays, effectively driving the actor loss towards zero. The critic loss, given in Eq.~\eqref{eq:critic_loss}, measures the difference between predicted and actual returns. The high initial value reflects the large scale of these returns within the warehouse environment: because agents receive a $R_\text{step} = -1$ penalty per step for non-productive actions and a $R_\text{pick} = +20$ reward for order picking, randomly initialized critics inherently produce substantial prediction errors. The loss decreases as the critic learns the value function.

\begin{figure}[!ht]
    \centering
    \begin{subfigure}{0.45\textwidth}
        \centering
        \includegraphics[width=\linewidth]{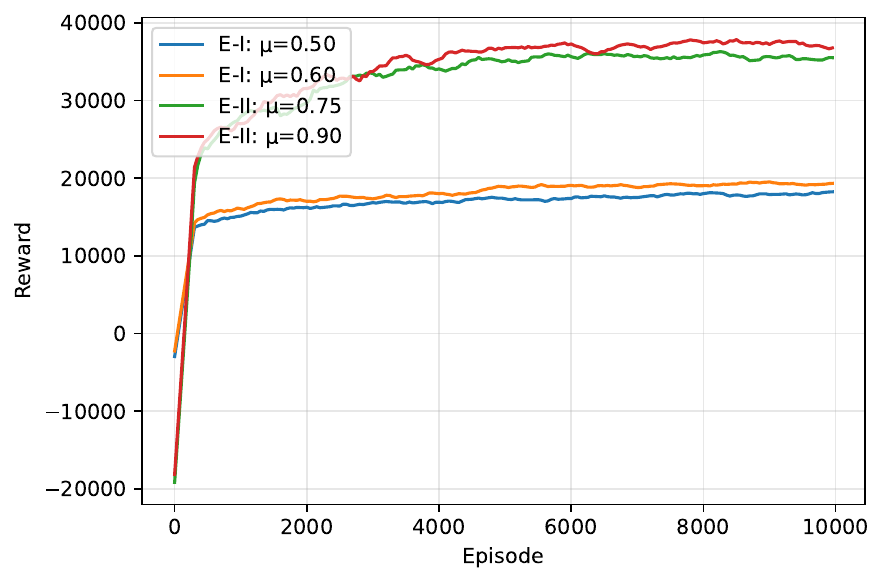}
        \caption{Reward}
    \end{subfigure}
    \hfill
    \begin{subfigure}{0.45\textwidth}
        \centering
        \includegraphics[width=\linewidth]{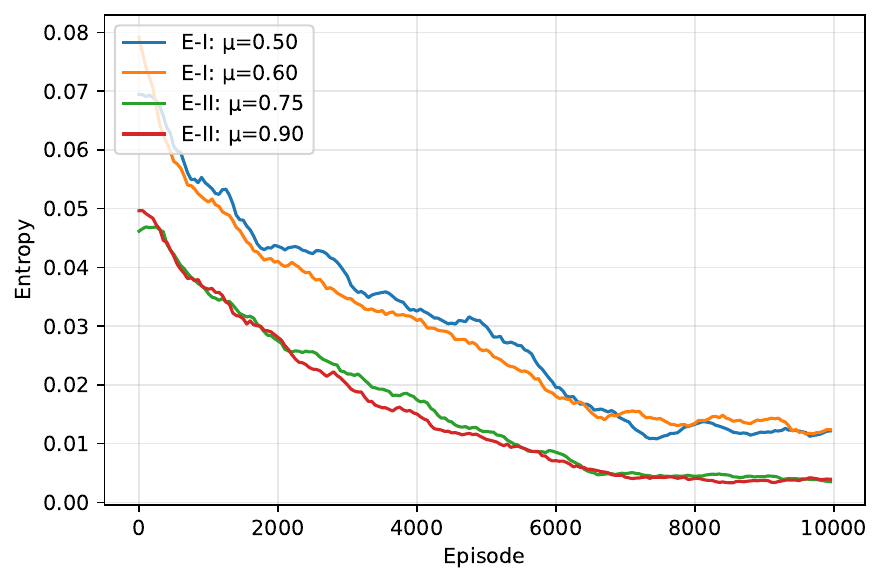}
        \caption{Entropy}
    \end{subfigure}
    \begin{subfigure}{0.45\textwidth}
        \centering
        \includegraphics[width=\linewidth]{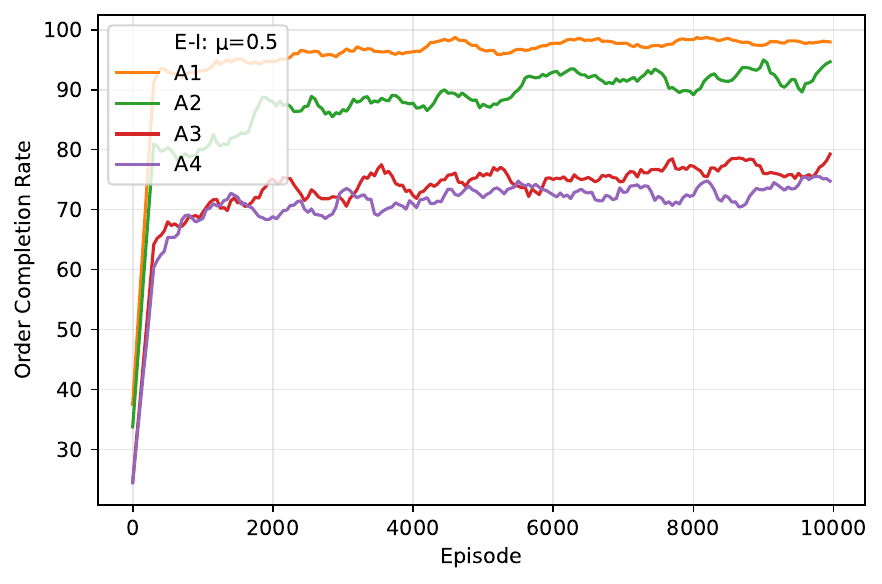}
        \caption{Order completion rate across agents in E-I environment with $\mu = 0.50$}
    \end{subfigure}
    \hfill
    \begin{subfigure}{0.45\textwidth}
        \centering
        \includegraphics[width=\linewidth]{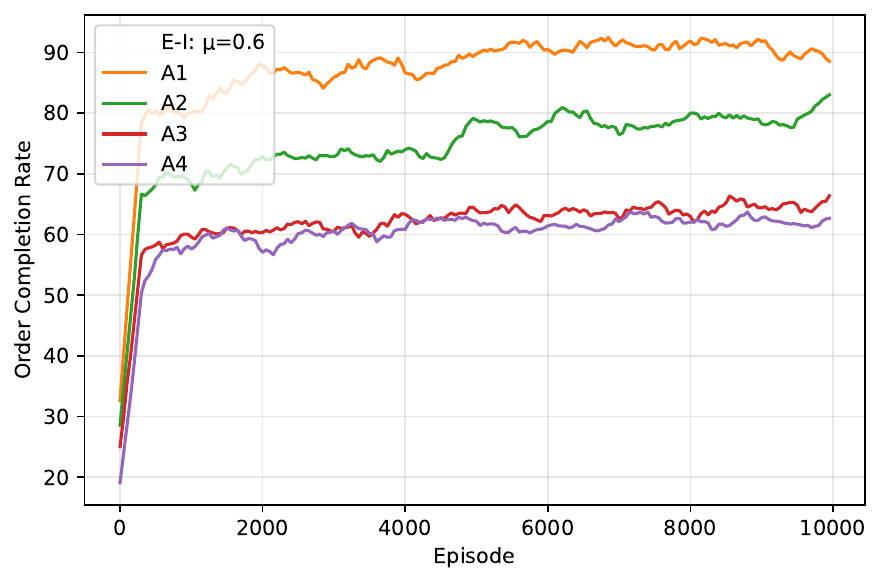}
        \caption{Order completion rate across agents in E-I environment with $\mu = 0.60$}
    \end{subfigure}
    \begin{subfigure}{0.45\textwidth}
        \centering
        \includegraphics[width=\linewidth]{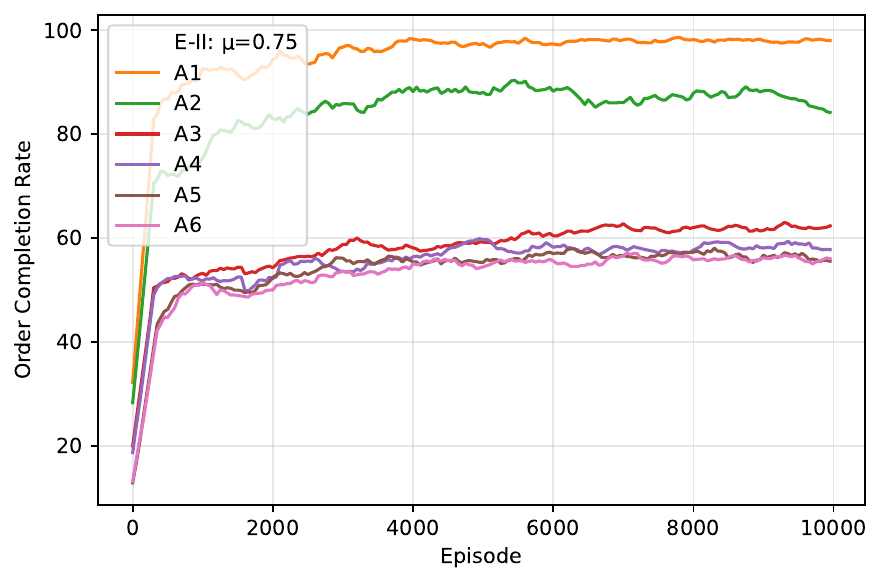}
        \caption{Order completion rate across agents in E-II environment with $\mu = 0.75$}
    \end{subfigure}
    \hfill
    \begin{subfigure}{0.45\textwidth}
        \centering
        \includegraphics[width=\linewidth]{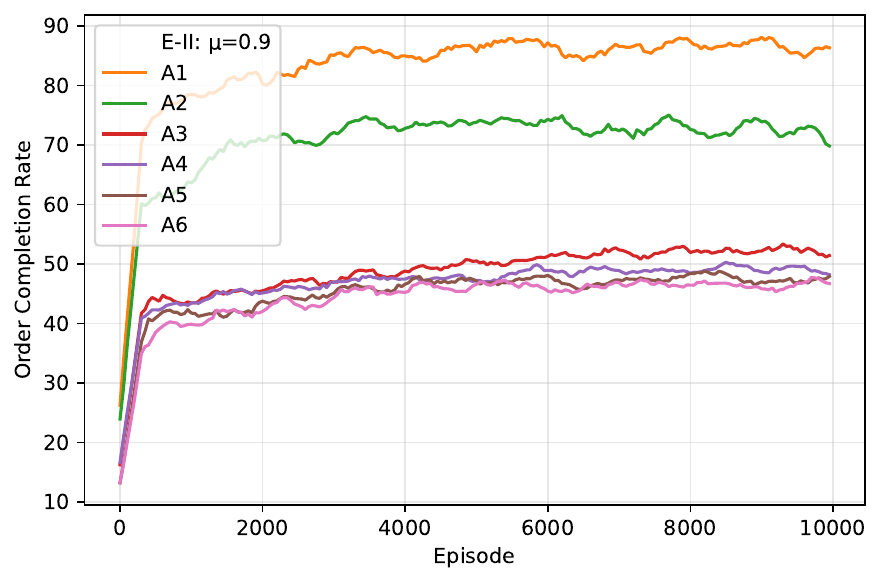}
        \caption{Order completion rate across agents in E-II environment with $\mu = 0.90$}
    \end{subfigure}
    \caption{Reward, entropy, and order completion rates under different scenarios.}
    \label{fig:s6_2}
\end{figure}

\begin{figure}[!ht]
    \centering
    \begin{subfigure}{0.45\textwidth}
        \centering
        \includegraphics[width=\linewidth]{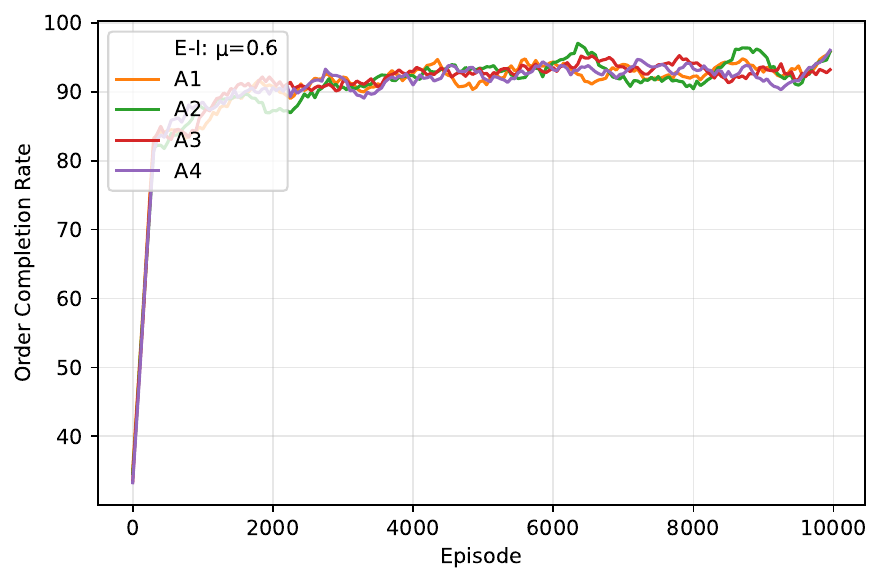}
    \end{subfigure}
    \caption{Order completion rate across agents in E-I environment with $\mu = 0.60$ and depot at the center}
    \label{fig:s6_5}
\end{figure}

The reward, entropy, and the number of orders completed per agent during training are shown in Figure~\ref{fig:s6_2}. The order completion rate varies between agents due to their different distances from the depot. Agents operating further from the depot spend a greater proportion of their time traveling to it, which reduces their overall order completion rate. We empirically verified this spatial dependency by relocating the depot to the center of the warehouse in E-I and retraining the model, which resulted in a uniform completion rate between agents as shown in Figure~\ref{fig:s6_5}.

%% file: tables/5_s6_env_hyper_params.tex
\begin{table*}[!ht]
\centering
\caption{Overview of experiment settings}
\label{tab:experiment_settings}
\setlength{\tabcolsep}{4.5pt}

\begin{subtable}[t]{0.48\textwidth}
    \centering
    \caption{Environment Configurations}
    \label{tab:env_parameters}
    \begin{tabular}{lcc}
    \toprule
    \textbf{Parameter} & \textbf{E-I} & \textbf{E-II} \\
    \midrule
    $N$                   & 4           & 6 \\
    $M$                   & 2           & 2 \\
    $N_{\text{aisle}}$    & 4           & 4 \\
    $N_{\text{slot}}$     & 8           & 8 \\
    $K$                   & 10          & 10 \\
    $b_{\max}$            & 100         & 100 \\
    $b_{\min}$            & 15          & 20 \\
    $\eta$                & 1           & 1 \\
    $\beta$               & 2           & 2 \\
    $V$                   & 1           & 1 \\
    $\gamma$              & 0.50, 0.60  & 0.75, 0.90 \\
    $\kappa^i$            & 1           & 1 \\
    \bottomrule
    \end{tabular}
\end{subtable}%
\hfill 
\begin{subtable}[t]{0.50\textwidth}
    \centering
    \caption{Model Hyperparameters}
    \label{tab:hyperparameters}
    \begin{tabular}{lcc}
    \toprule
    \textbf{Parameter} & \textbf{Search Range} & \textbf{Value} \\
    \midrule
    \multicolumn{3}{l}{\textit{Optimiser}} \\
    $\text{lr}_{\text{actor}}$        & $[10^{-4},\ 5\times10^{-4}]$  & $1.20 \times 10^{-4}$ \\
    $\text{lr}_{\text{critic}}$       & $[10^{-4},\ 10^{-3}]$         & $4.86 \times 10^{-4}$ \\
    \hline 
    \multicolumn{3}{l}{\textit{PPO \& GAE}} \\ 
    $N_{\text{actor\_update}}$        & $\{1, 2, 3, 4\}$              & 2 \\
    $N_{\text{critic\_update}}$       & $\{1,\ldots,10\}$             & 3 \\
    Mini-batch size                   & -                             & 512 \\
    Gradient clip                     & -                             & 0.5 \\
    $\varepsilon$                     & $[0.1,\ 0.3]$                 & 0.208 \\
    $\lambda$                         & $[0.90,\ 0.99]$               & 0.98 \\
    $\alpha_{\bar{r}}$                & $[0.01,\ 0.1]$                & 0.0117 \\
    $H_{0}$                           & $[0.05,\ 0.20]$               & 0.164 \\
    \hline
    \multicolumn{3}{l}{\textit{Training}} \\
    Episodes $E$                & -                             & 10000 \\
    $T$                               & -                             & 14400 s \\
    \bottomrule
    \end{tabular}
\end{subtable}
\end{table*}

%% file: sections/8_benchmarking_PPO.tex
\subsection{Benchmarking}
As discussed in section~2, DQN is a prevalent and highly effective MARL algorithm in warehouse operations. To evaluate our approach, we benchmark our model against a DQN baseline and a centralized training decentralized execution (CTDE) PPO model.

To benchmark against existing research, we adopt the DRL framework implemented by \citet{bischoff2025reinforcement}. The study considers a single block warehouse where AMRs make a charging decision post completion of transport tasks which can be retrieval or delivery. The state space comprises scaled battery level, distance to the charging stations, the number of pending charging events and a lane-wise entropy. The action specifies the battery level up to which an agent charges. Actions correspond to either not charging (action = 0) or charging up to a specified battery level. The charging threshold is a design choice where two options are proposed. \textit{$A\_\text{full}$}, where the agent can choose action 0 or an integer multiple of 10 and \textit{$A\_\text{binary}$} where it can choose action 0 or 100. Action masks are used to prevent illegal actions. A minimum battery threshold is enforced, below which charging is mandatory. Multiple reward functions are proposed that consider service time, queued orders, availability of free AMRs, explicit penalties and incentives for charging decisions. An interrupt strategy is further employed, whereby if retrieval orders are pending and no AMR is available, charging for any AMR whose battery level exceeds 50\% is halted. For re-charging, the nearest CS is given the first preference, followed by the CS with shortest queue length. The framework employs the masked PPO algorithm from \citet{SB3}.

We adapt their proposed framework to a multi-block environment by including only retrieval operations and introducing dynamic order arrivals. We utilize the same masked PPO network, number of training steps, buffer size, batch size and network parameters, action space, action masks and CS selection rules. Consistent with their framework, the state space includes the number of AMRs in different conditions, such as free to move, charging, and idle; the battery levels of working and charging AMRs, distance to CS, queue length at CS, and time. We also adopt their \textit{shaped} reward function: $+1$ if the agent chooses to charge while free CSs are available, and $-1$ otherwise. Similarly, charging when retrieval orders are pending yields a reward of $-1$, otherwise $+1$. Depot travel is restricted to instances when carrying capacity is fully exhausted. For this experiment, the best results were generated with the interrupt strategy disabled and \textit{$A\_\text{full}$} as the battery charging design. The framework is further compared with heuristic strategies such as the \textit{FixedThreshold} strategy where a lower and upper threshold for the battery is fixed and \textit{HighLow} strategy where the upper threshold is varied depending on the retrieval order queue length. In our implementation, this quantity is represented by the percentage of remaining orders.

\begin{figure*}[!ht]
    \centering
    \begin{subfigure}{0.45\textwidth}
        \centering
        \includegraphics[width=\linewidth]{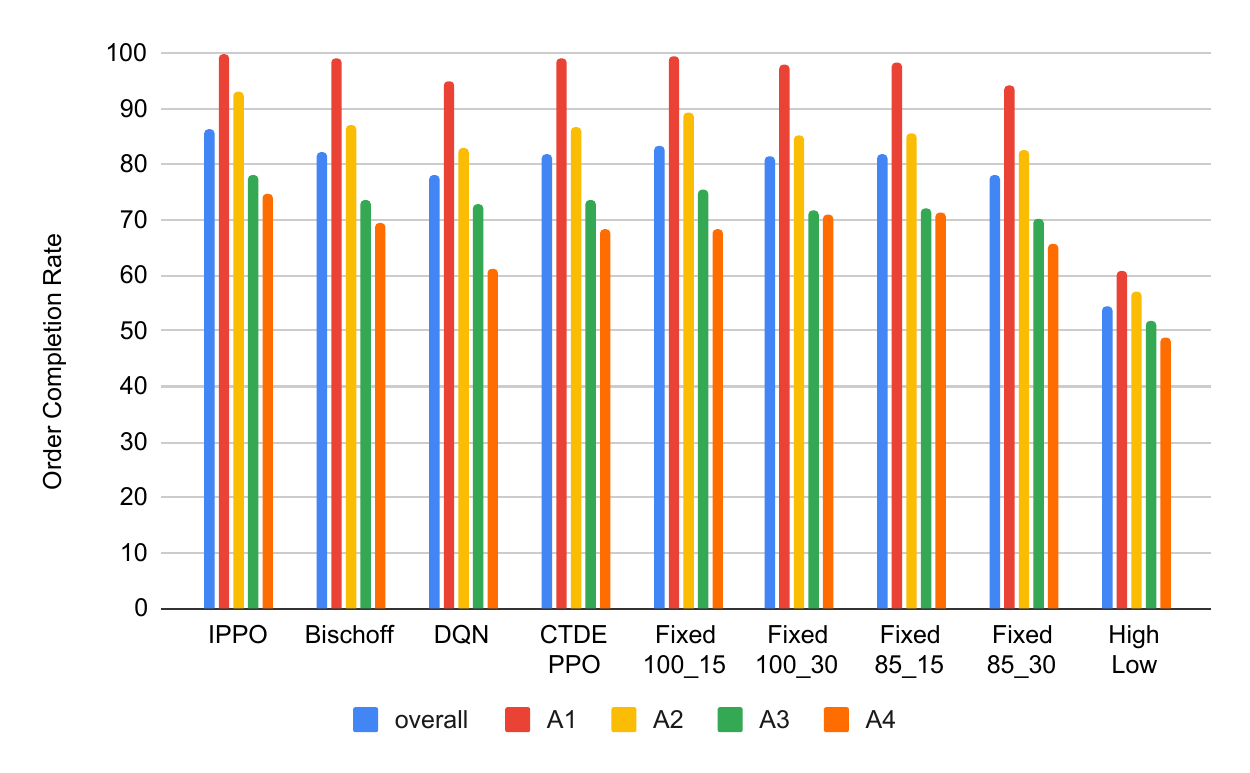}
        \caption{E-I environment with $\mu$ = 0.50}
    \end{subfigure}
    \hfill
    \begin{subfigure}{0.45\textwidth}
        \centering
        \includegraphics[width=\linewidth]{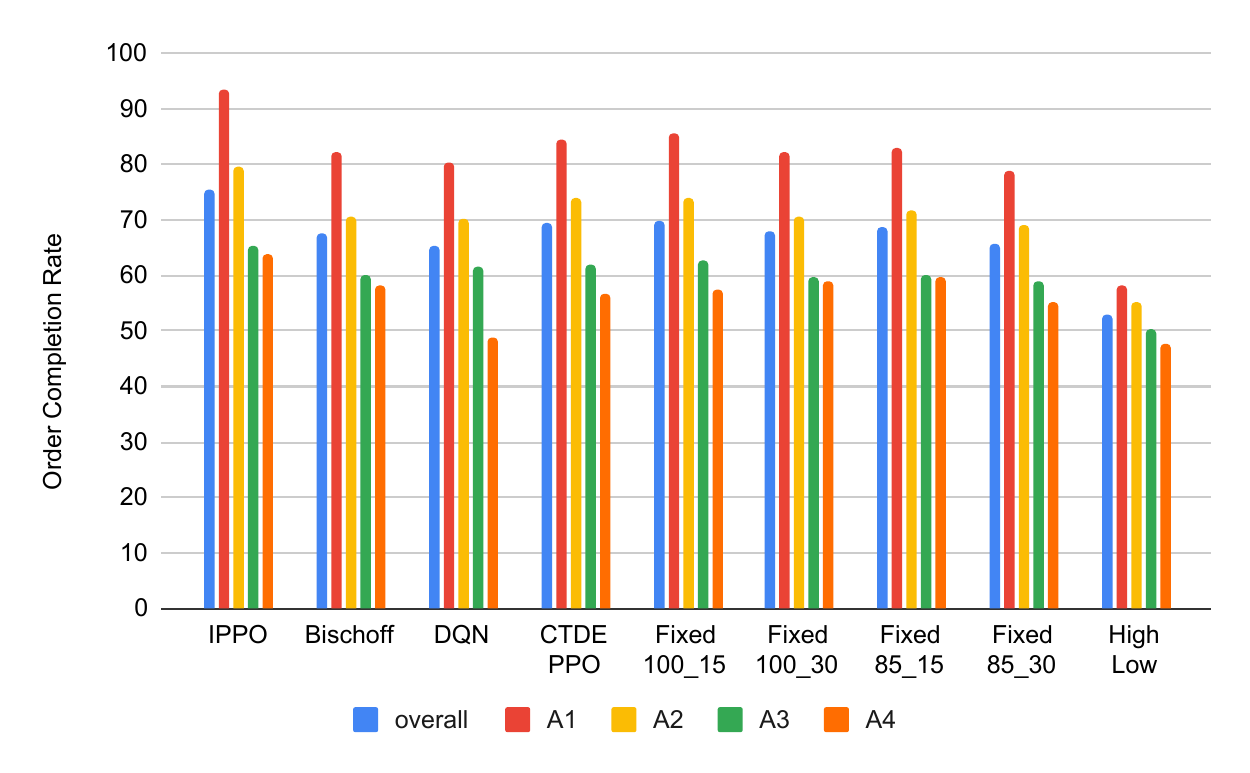}
        \caption{E-I environment with $\mu$ = 0.60}
    \end{subfigure}
    \begin{subfigure}{0.45\textwidth}
        \centering
        \includegraphics[width=\linewidth]{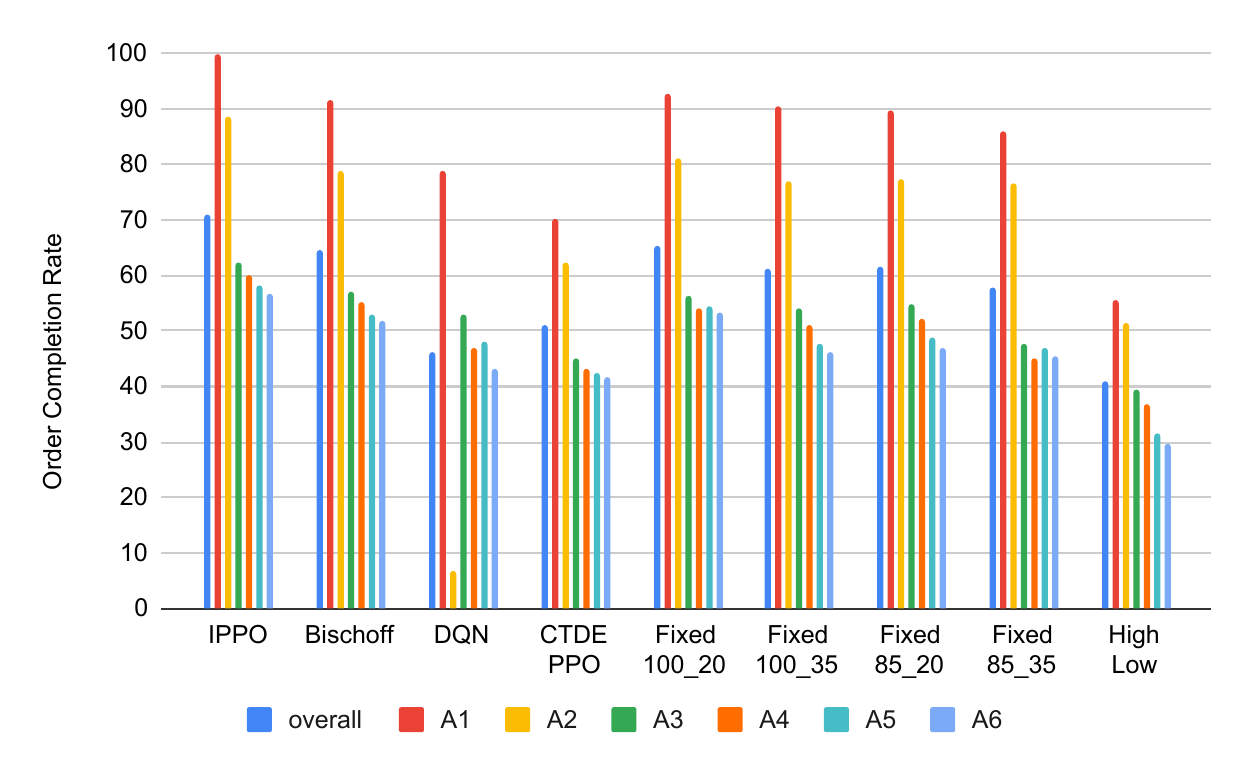}
        \caption{E-II environment with $\mu$ = 0.75}
    \end{subfigure}
    \hfill
    \begin{subfigure}{0.45\textwidth}
        \centering
        \includegraphics[width=\linewidth]{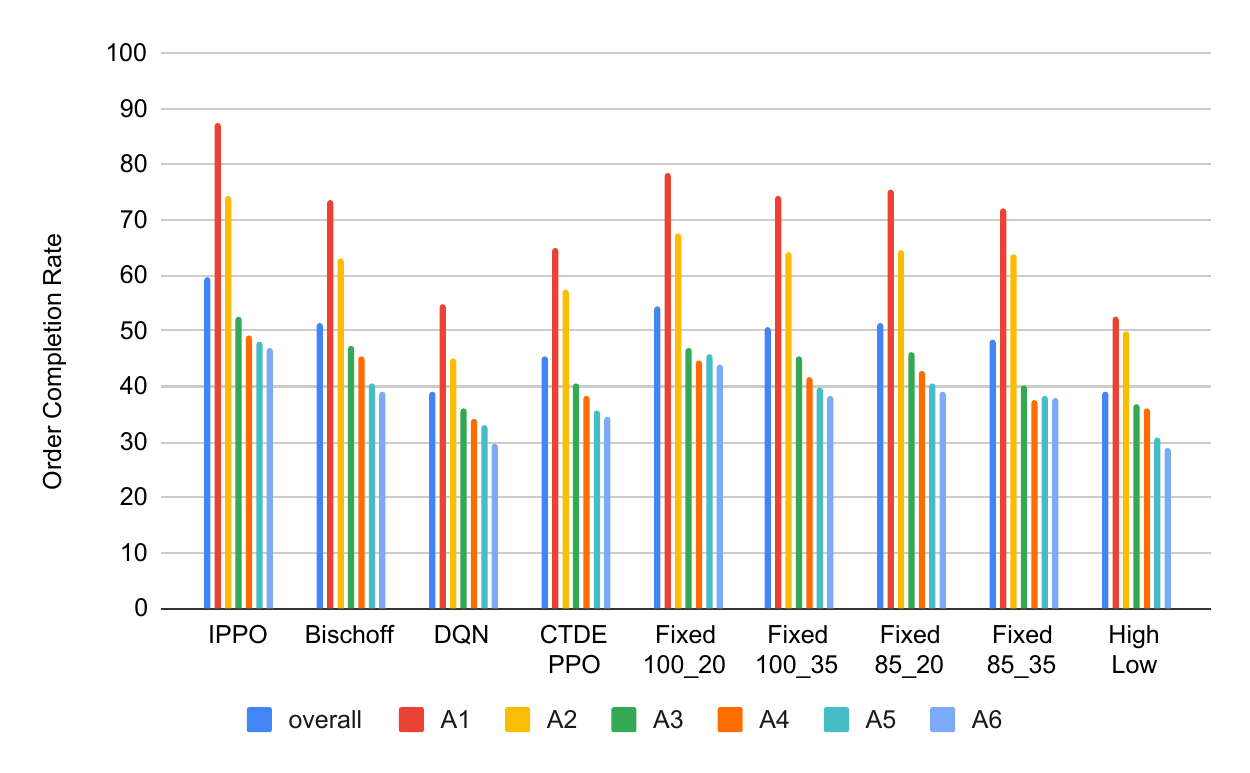}
        \caption{E-II environment with $\mu$ = 0.90}
    \end{subfigure}    
    \caption{Performance across different benchmarking algorithms}
    \label{fig:s8_1}
\end{figure*}

IPPO outperformed all benchmark algorithms as shown in Figure~\ref{fig:s8_1}, Table~\ref{tab:order_completion_EI}, and Table~\ref{tab:order_completion_EII}. As the complexity increases ($\mu = 0.6$), IPPO achieves a higher order completion rate than the best performing benchmark. Among the benchmarking algorithms, \textit{FixedThreshold} achieves the best performance when the lower threshold is minimized and the upper threshold is maximized. For example, in E-I with $\mu = 0.50$, \textit{Fixed 100\_15} which corresponds to \textit{FixedThreshold} strategy with lower threshold $15$ and upper threshold $100$ is the best performing benchmark.

\input{tables/10_s8_results_EI}
\input{tables/11_s8_results_EII}

\begin{figure*}[!ht]
    \centering
    \begin{subfigure}{0.45\textwidth}
        \centering
        \includegraphics[width=\linewidth]{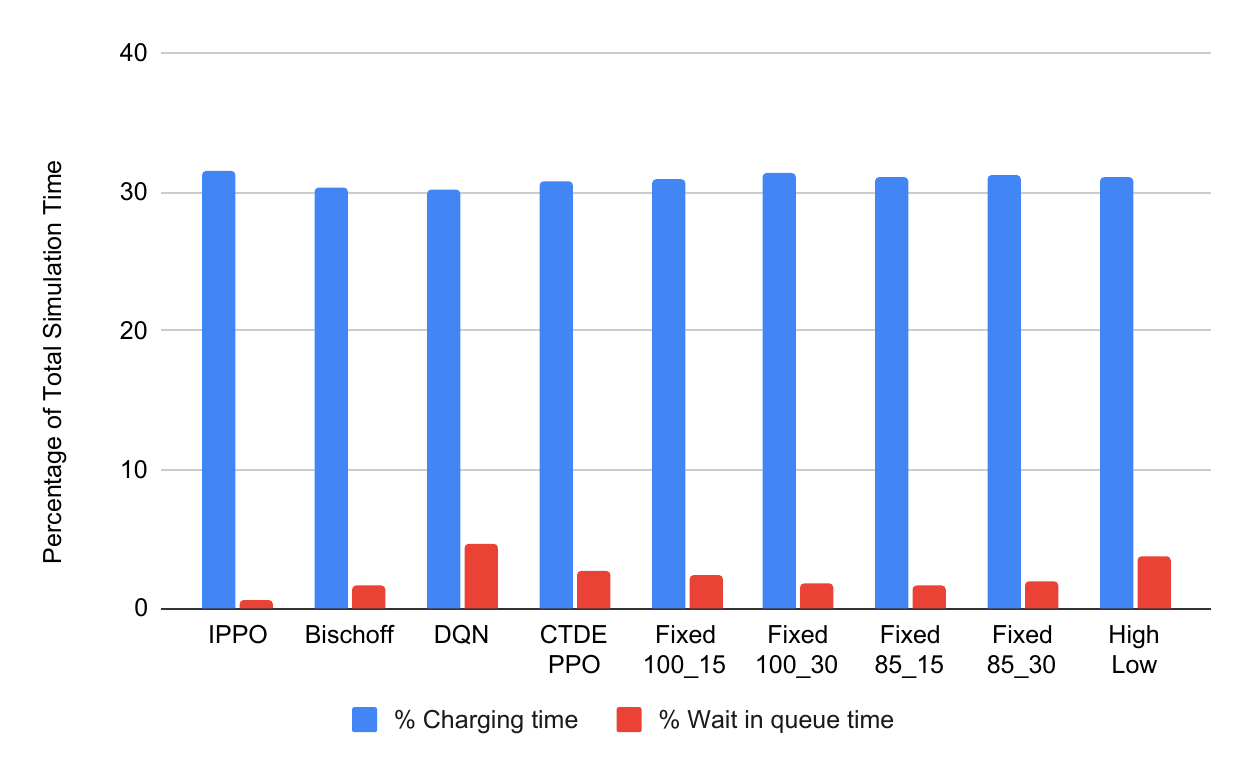}
        \caption{E-I environment with $\mu$ = 0.60}
    \end{subfigure}
    \hfill
    \begin{subfigure}{0.45\textwidth}
        \centering
        \includegraphics[width=\linewidth]{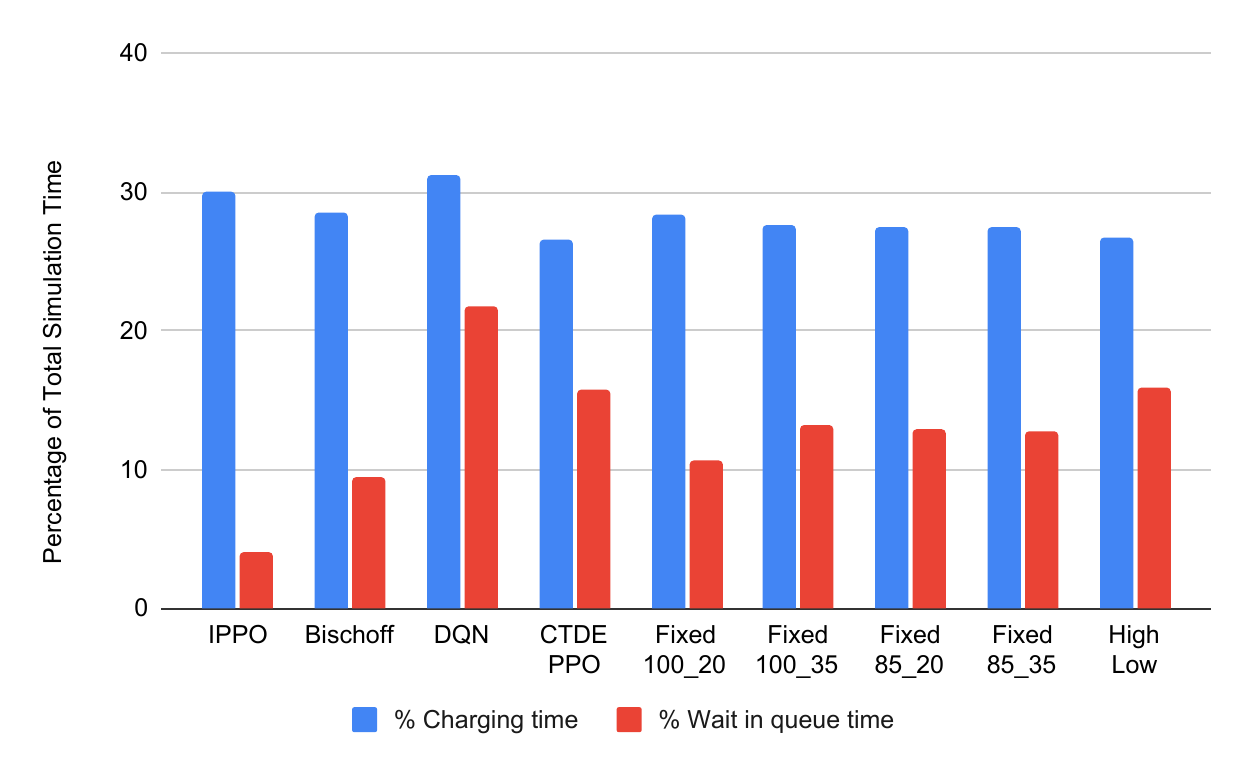}
        \caption{E-II environment with $\mu$ = 0.90}
    \end{subfigure}
    \caption{Percentage of time spent charging and waiting in queue at CS in different scenarios}
    \label{fig:s8_2}
\end{figure*}

\begin{figure*}[!ht]
    \centering
    \begin{subfigure}{0.45\textwidth}
        \centering
        \includegraphics[width=\linewidth]{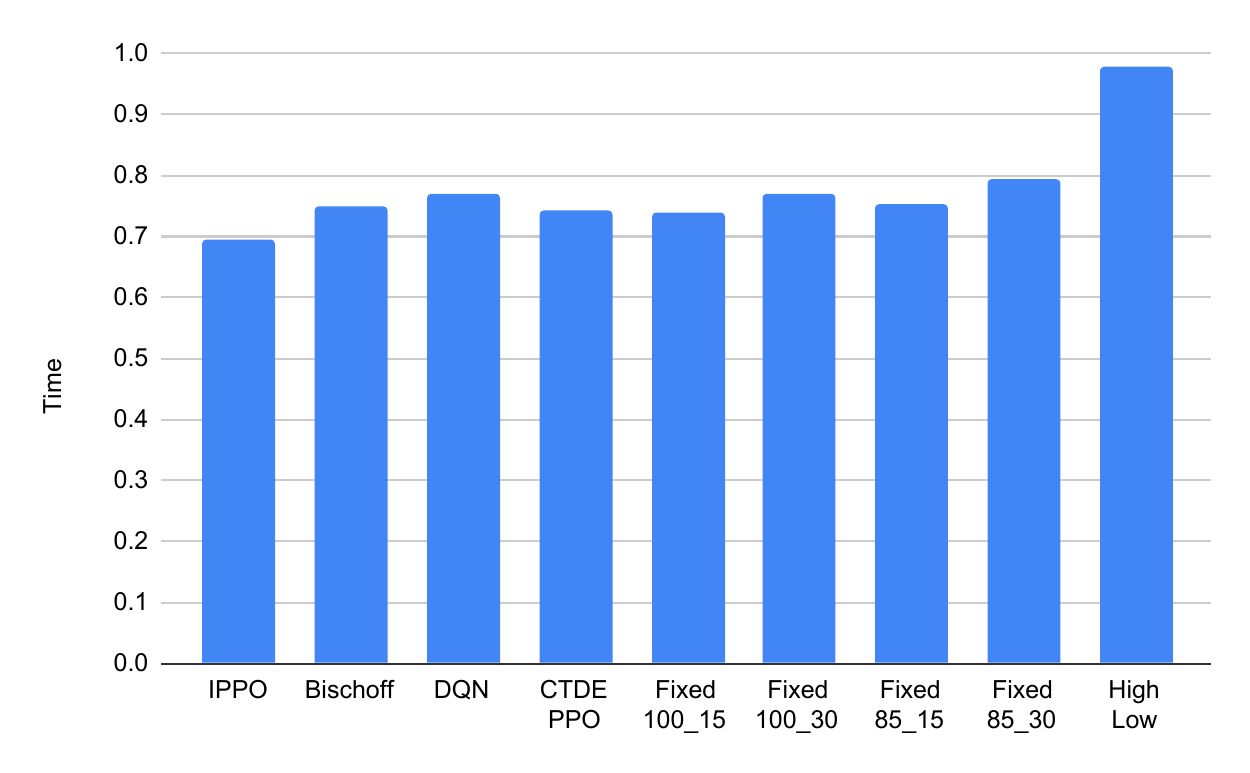}
        \caption{E-I environment with $\mu$ = 0.60}
    \end{subfigure}
    \hfill
    \begin{subfigure}{0.45\textwidth}
        \centering
        \includegraphics[width=\linewidth]{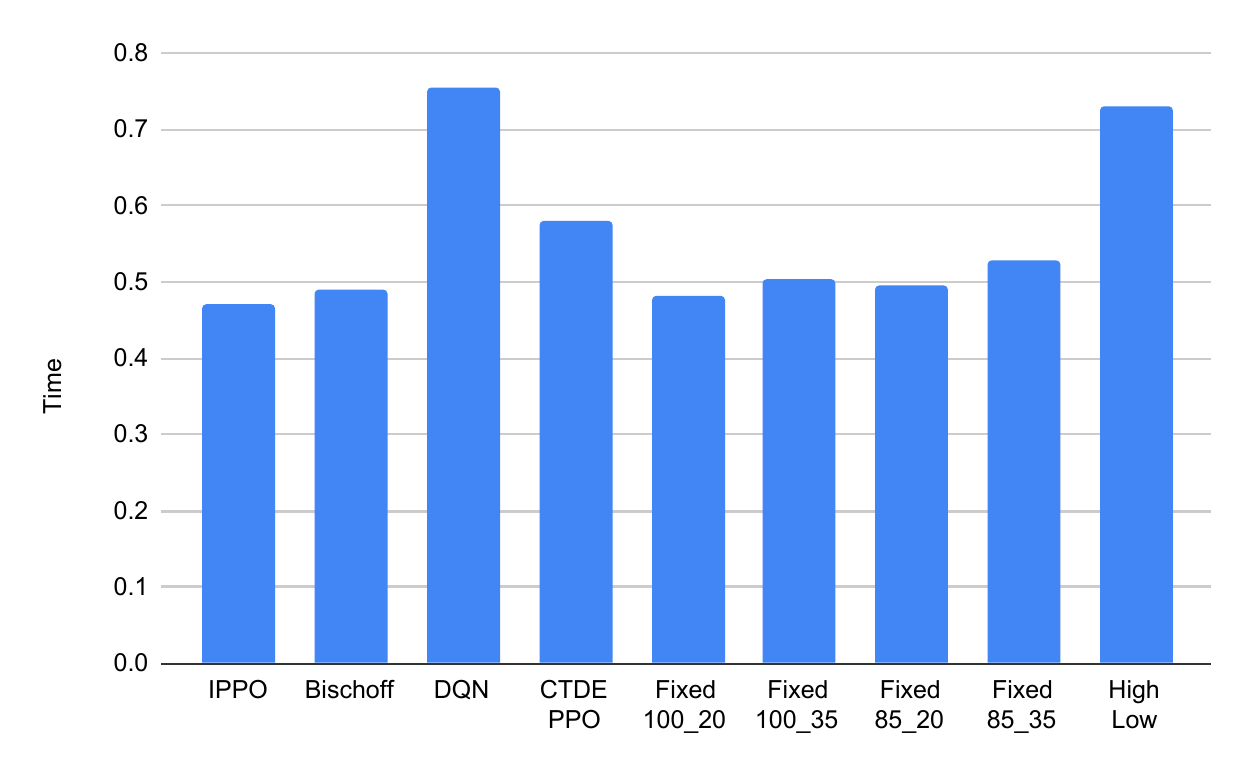}
        \caption{E-II environment with $\mu$ = 0.90}
    \end{subfigure}
    \caption{Time spent charging for per completed order in different scenarios}
    \label{fig:s8_3}
\end{figure*}

Among all benchmark algorithms, IPPO achieves the lowest average waiting time at charging stations and lowest time spent charging per order completed as shown in Figures ~\ref{fig:s8_2} and ~\ref{fig:s8_3}.

%% file: tables/10_s8_results_EI.tex
\begin{table*}[!ht]
\centering
\caption{Order Completion Rate (\%) under different algorithms in E-I settings}
\label{tab:order_completion_EI}
\resizebox{\textwidth}{!}{%
\begin{tabular}{llccccccccc}
\toprule
$\mu$ & Algo. & IPPO & Bischoff & DQN & CTDE PPO & Fixed 100\_15 & Fixed 100\_30 & Fixed 85\_15 & Fixed 85\_30 & High Low \\
\midrule
\multirow{5}{*}{0.6}
& overall & \textbf{76} & 68 & 65 & 69 & \textbf{70} & 68 & 69 & 66 & 53 \\
& A1 & 94 & 82 & 80 & 85 & 85 & 82 & 83 & 79 & 58 \\
& A2 & 80 & 71 & 70 & 74 & 74 & 71 & 72 & 69 & 55 \\
& A3 & 66 & 60 & 61 & 62 & 63 & 60 & 60 & 59 & 50 \\
& A4 & 64 & 58 & 49 & 57 & 57 & 59 & 60 & 55 & 48 \\
\midrule
\multirow{5}{*}{0.5}
& overall & \textbf{87} & 82 & 78 & 82 & \textbf{83} & 82 & 82 & 78 & 55 \\
& A1 & 100 & 99 & 95 & 99 & 100 & 98 & 99 & 94 & 61 \\
& A2 & 93 & 87 & 83 & 87 & 89 & 85 & 86 & 83 & 57 \\
& A3 & 78 & 73 & 73 & 73 & 76 & 72 & 72 & 70 & 52 \\
& A4 & 75 & 69 & 61 & 68 & 69 & 71 & 71 & 66 & 49 \\
\bottomrule
\end{tabular}}
\end{table*}

%% file: tables/11_s8_results_EII.tex
\begin{table*}[ht]
\centering
\caption{Order Completion Rate (\%) under different algorithms in E-II settings}
\label{tab:order_completion_EII}
\resizebox{\textwidth}{!}{%
\begin{tabular}{llccccccccc}
\hline
$\mu$ & Algo. & IPPO & Bischoff & DQN & CTDE PPO & Fixed 100\_20 & Fixed 100\_35 & Fixed 85\_20 & Fixed 85\_35 & High Low \\
\hline
\multirow{7}{*}{0.9}
& overall & \textbf{60} & 52 & 39 & 45 & \textbf{55} & 51 & 51 & 48 & 39 \\
& A1 & 87 & 74 & 55 & 65 & 78 & 74 & 75 & 72 & 52 \\
& A2 & 74 & 63 & 45 & 57 & 68 & 64 & 65 & 64 & 50 \\
& A3 & 53 & 47 & 36 & 40 & 47 & 46 & 46 & 40 & 37 \\
& A4 & 49 & 45 & 34 & 38 & 45 & 42 & 43 & 38 & 36 \\
& A5 & 48 & 41 & 33 & 36 & 46 & 40 & 41 & 38 & 31 \\
& A6 & 47 & 39 & 30 & 35 & 44 & 38 & 39 & 38 & 29 \\
\hline
\multirow{7}{*}{0.75}
& overall & \textbf{71} & 65 & 46 & 51 & \textbf{65} & 61 & 62 & 58 & 41 \\
& A1 & 100 & 91 & 79 & 70 & 93 & 91 & 90 & 86 & 56 \\
& A2 & 89 & 79 & 7 & 62 & 81 & 77 & 77 & 76 & 52 \\
& A3 & 62 & 57 & 53 & 45 & 56 & 54 & 55 & 48 & 39 \\
& A4 & 60 & 55 & 47 & 43 & 54 & 51 & 52 & 45 & 37 \\
& A5 & 58 & 53 & 48 & 42 & 54 & 48 & 49 & 47 & 31 \\
& A6 & 57 & 52 & 43 & 42 & 53 & 46 & 47 & 46 & 30 \\
\hline
\end{tabular}}
\end{table*}

%% file: sections/7_explainability.tex
\subsection{Explainability and Policy Analysis}

\begin{figure*}[!ht]
    \centering
    \begin{subfigure}{0.45\textwidth}
        \centering
        \includegraphics[width=\linewidth]{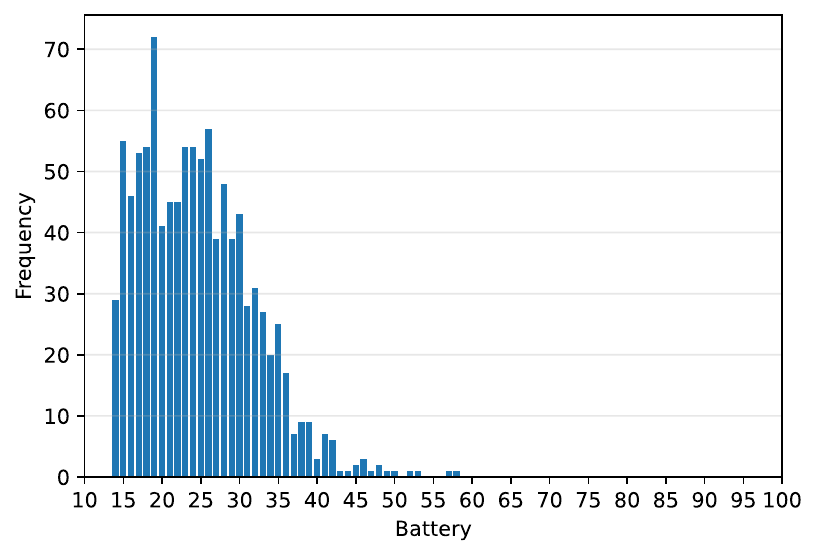}
        \caption{Battery when agents decide to go to CS}
    \end{subfigure}
    \hfill
    \begin{subfigure}{0.45\textwidth}
        \centering
        \includegraphics[width=\linewidth]{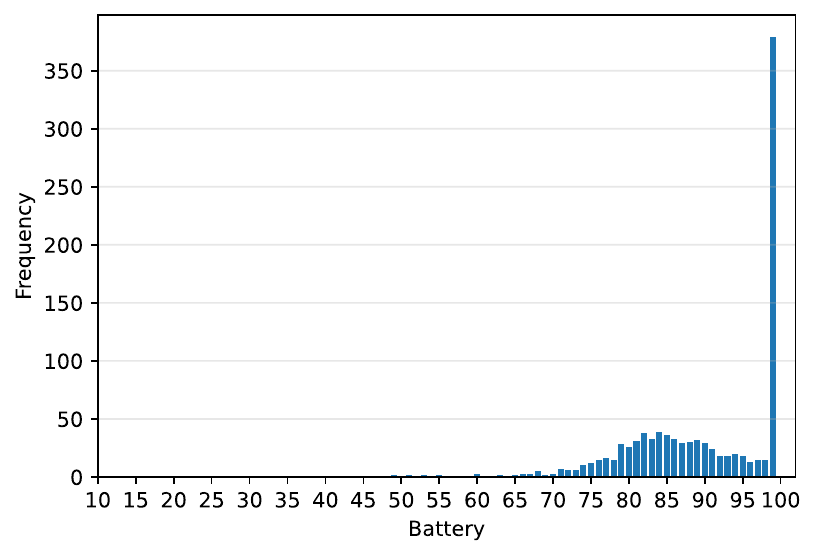}
        \caption{Battery when agents decide to stop charging}
    \end{subfigure}
    \begin{subfigure}{0.45\textwidth}
        \centering
        \includegraphics[width=\linewidth]{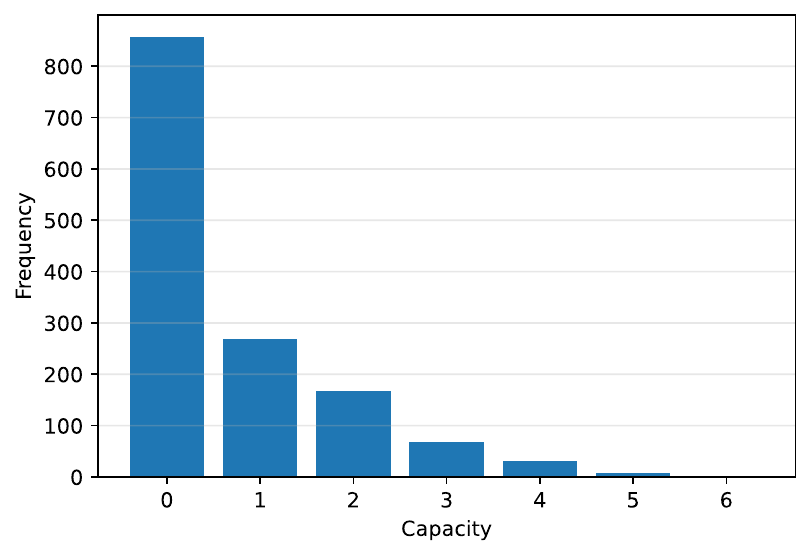}
        \caption{Capacity when agents decide to go to depot}
    \end{subfigure}
    \hfill
    \begin{subfigure}{0.45\textwidth}
        \centering
        \includegraphics[width=\linewidth]{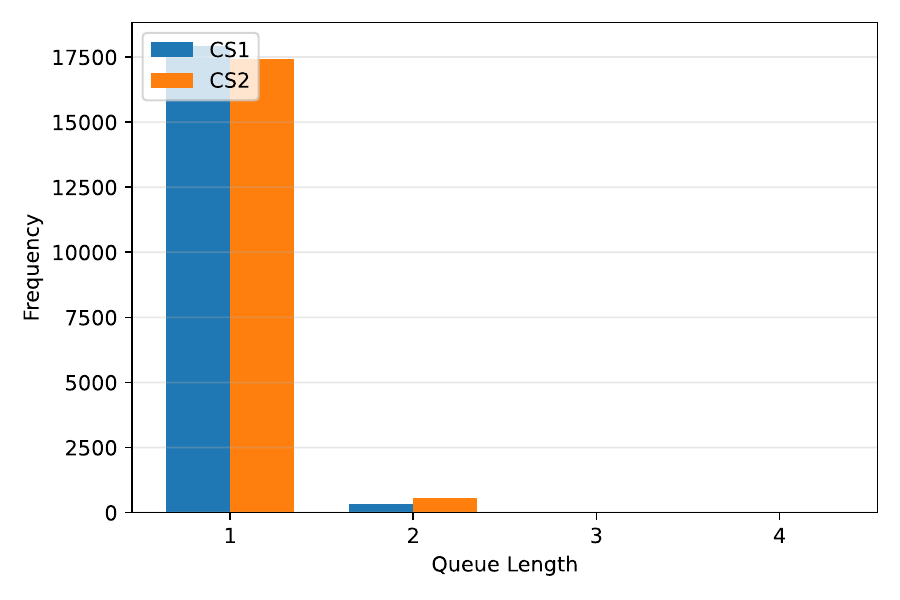}
        \caption{Queue length at CS}
    \end{subfigure}
    \caption{Policy learned in E-I environment with $\mu = 0.60$}
    \label{fig:s6_3}
\end{figure*}

\begin{figure*}[!ht]
    \centering
    \begin{subfigure}{0.45\textwidth}
        \centering
        \includegraphics[width=\linewidth]{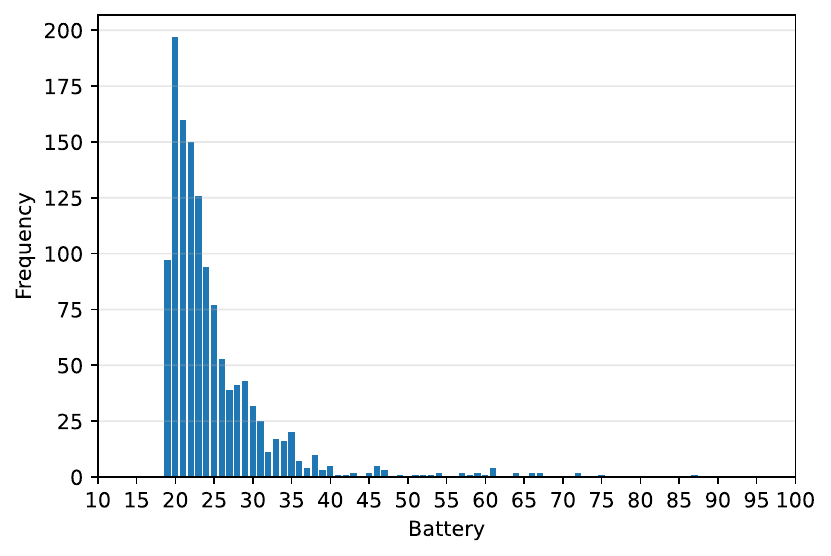}
        \caption{Battery when agents decide to go to CS}
    \end{subfigure}
    \hfill
    \begin{subfigure}{0.45\textwidth}
        \centering
        \includegraphics[width=\linewidth]{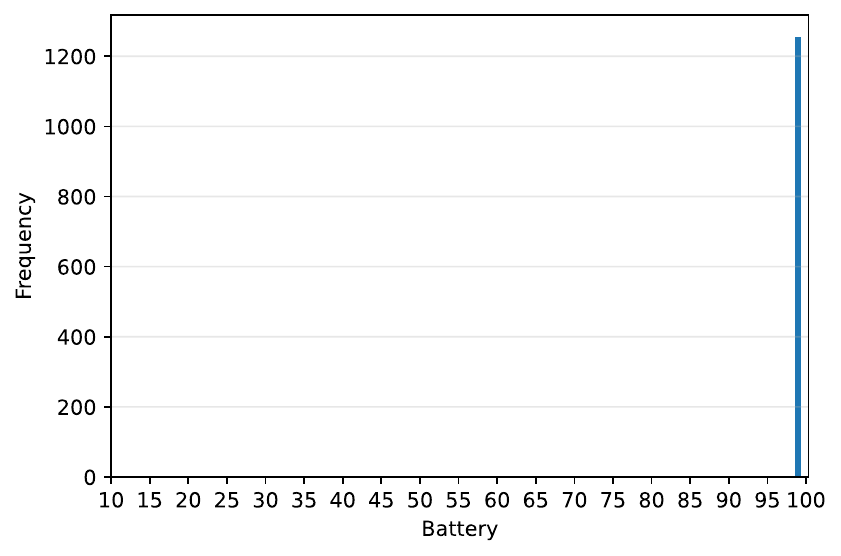}
        \caption{Battery when agents decide to stop charging}
    \end{subfigure}
    \begin{subfigure}{0.45\textwidth}
        \centering
        \includegraphics[width=\linewidth]{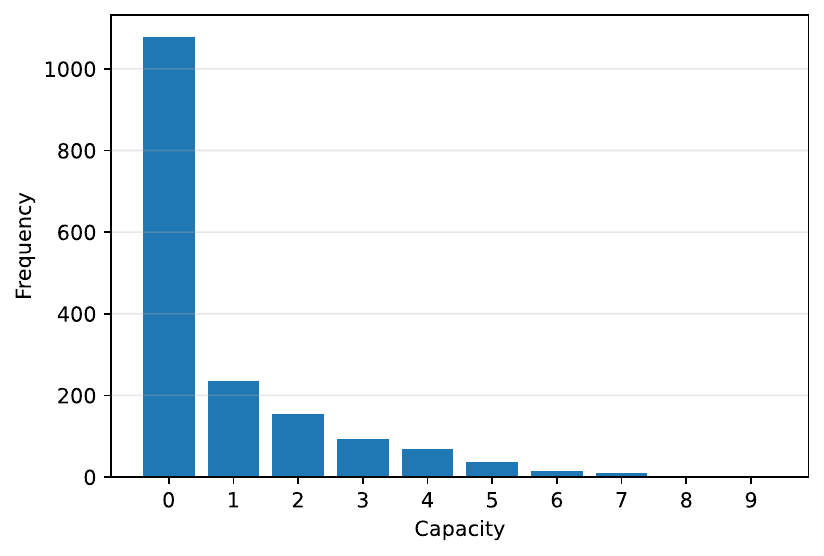}
        \caption{Capacity when agents decide to go to depot}
    \end{subfigure}
    \hfill
    \begin{subfigure}{0.45\textwidth}
        \centering
        \includegraphics[width=\linewidth]{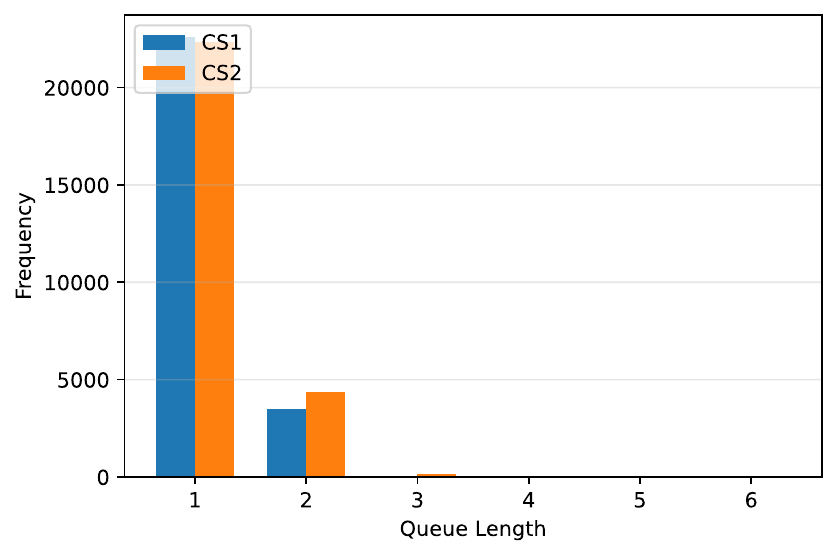}
        \caption{Queue length at CS}
    \end{subfigure}
    \caption{Policy learnt in E-II environment with $\mu = 0.90$}
    \label{fig:s6_4}
\end{figure*}

To characterize the policies learned by the model, the battery level at which agents make charging decisions, capacity level when they go to depot and queue length at CS are analyzed in environments E-I and E-II with $\mu = 0.6$ and $\mu = 0.9$ respectively. Figures~\ref{fig:s6_3} and ~\ref{fig:s6_4} present the distributions of battery levels at charging decisions, capacity level at deport decisions, and the queue lengths at charging stations. In both environments, agents most frequently go to the depot when their carrying capacity reaches its minimum and initiate charging when the battery level is low. In E-II, agents predominantly terminate charging after reaching full battery capacity, whereas in E-I they frequently stop charging at battery levels between 75\% and 95\%. The low frequency of longer queue lengths indicates that the learned policy effectively mitigates congestion at CS.

\subsubsection{SHAP Value Analysis}
A policy analysis is conducted using SHapley Additive exPlanations (SHAP) \citep{lundberg2017shap}. SHAP identifies the state features in the observation space that most strongly influence the agent's action selection. It measures the change in action probability assigned by the actor network when a feature is absent, averaged over all possible feature combinations. Features that consistently influence the prediction receive larger SHAP values, whereas features with little influence receive values close to zero.

\begin{figure*}[!ht]
    \centering
    \begin{subfigure}{0.9\textwidth}
        \centering
        \includegraphics[width=\linewidth]{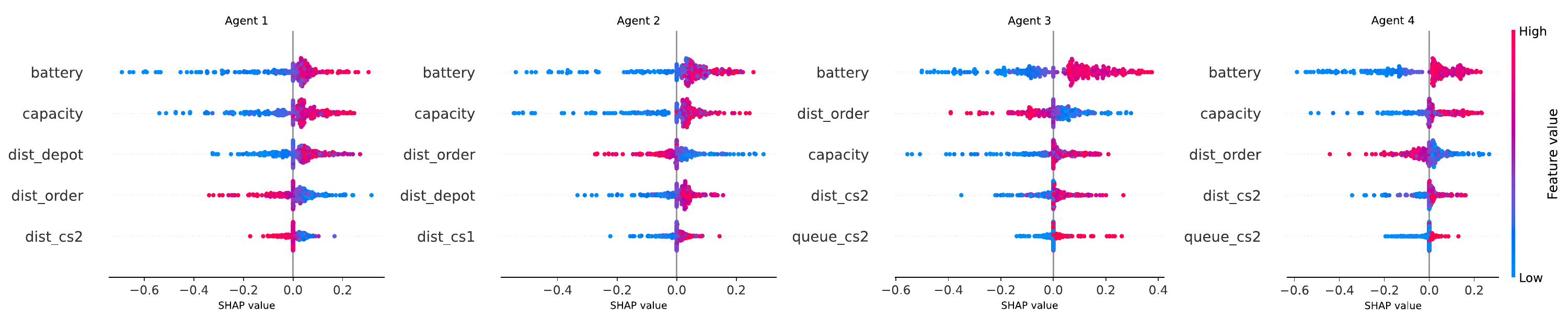}
        \caption{SHAP values for action go\_to\_pick}
    \end{subfigure}%
    \vspace{0.01cm}
    \begin{subfigure}{0.9\textwidth}
        \centering
        \includegraphics[width=\linewidth]{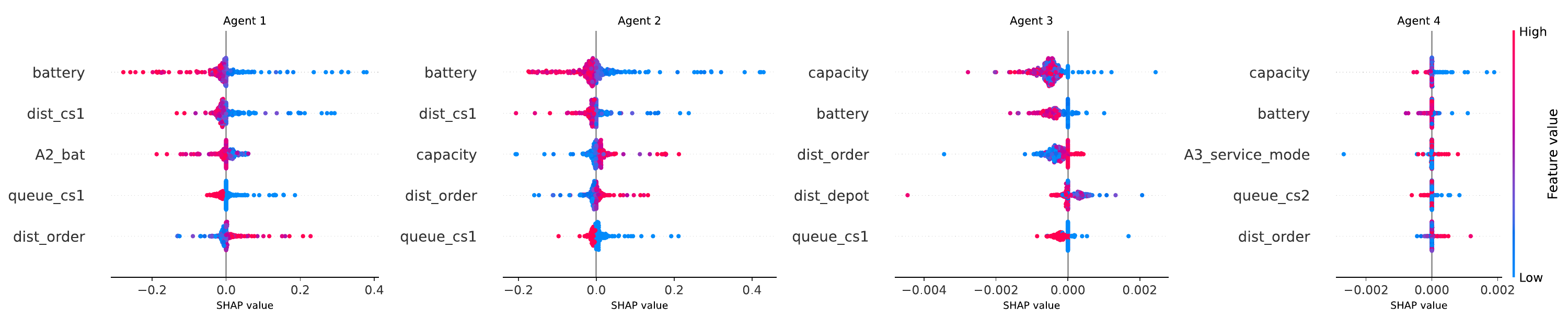}
        \caption{SHAP values for action go\_to\_CS1}
    \end{subfigure}%
    \vspace{0.01cm}
    \begin{subfigure}{0.9\textwidth}
        \centering
        \includegraphics[width=\linewidth]{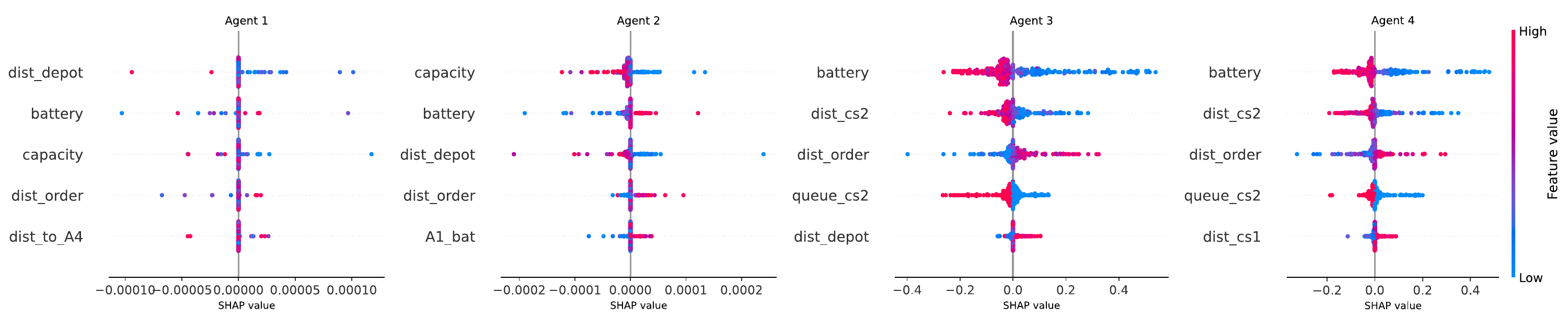}
        \caption{SHAP values for action go\_to\_CS2}
    \end{subfigure}%
    \vspace{0.01cm}
    \begin{subfigure}{0.9\textwidth}
        \centering
        \includegraphics[width=\linewidth]{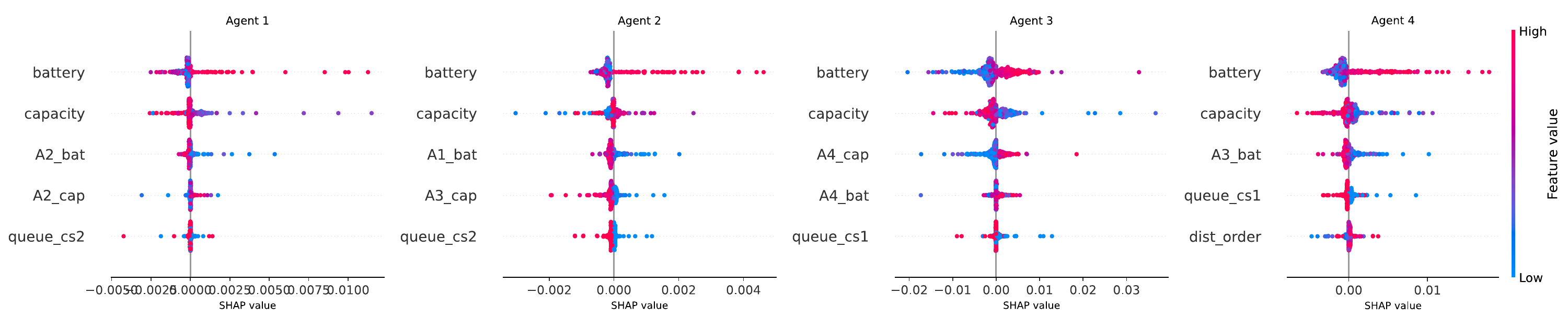}
        \caption{SHAP values for action stop\_charging}
    \end{subfigure}
    \vspace{0.01cm}
    \begin{subfigure}{0.9\textwidth}
        \centering
        \includegraphics[width=\linewidth]{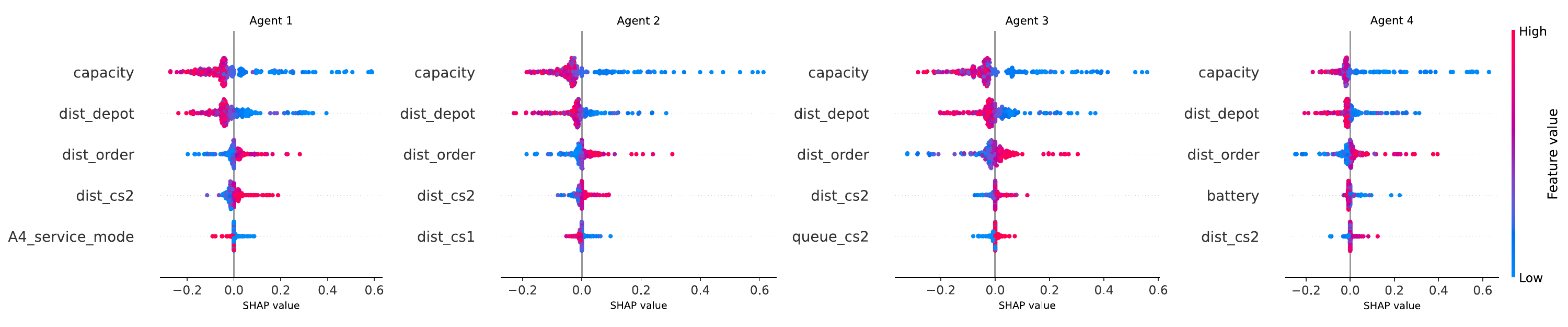}
        \caption{SHAP values for action go\_to\_depot}
    \end{subfigure}
    \caption{SHAP values for different actions based on policy learned for E-I environment with $\mu = 0.60$. Each dot in the beeswarm plot represents one decision instance. The horizontal position gives the SHAP value: dots to the right indicate this feature increased the probability of the action in that instance; dots to the left indicate it decreased it. The color encodes the raw feature value: red indicates a high feature value (e.g. high battery), blue indicates a low value (e.g. low battery).}
    \label{fig:s7_1}
\end{figure*}

Decision states are collected from 10 test episodes for both the E-I ($\mu = 0.6$) and E-II ($\mu = 0.9$) environments, retaining only time steps where the action mask permits more than one valid action to filter out mid-travel states. Further restrictions are placed since certain actions are available only under specific environmental conditions. For example, \texttt{stop\_charging} is available only when the agent is at a CS and first in the queue. To characterize the average policy behavior, 100 state observations are sampled as the background reference and used to explain 300 state observations. For representation, the five most influential features, ranked by their mean absolute SHAP value are reported in Figure~\ref{fig:s7_1} for E-I environment.

Based on the SHAP values, we can summarize the learned policy for all actions as follows:
\begin{itemize}[itemsep=0.0pt, topsep=0.0pt]
\item \texttt{go\_pick:} Battery, capacity and distance to the next order are the most influential. Red dots for battery appear on the right, confirming that agents pick orders when battery is sufficient, whereas blue dots for distance to the next order appear on the right, indicating that agents prefer picking when the order is close. Features related to neighboring agents have near-zero SHAP values, indicating picking decisions are primarily self-interested.
\item \texttt{go\_to\_CS:} Battery is the most influential feature, with low battery levels strongly promoting the charging decision. Queue lengths at CS also influences the decision. Agents exhibit a preference for the nearest CS. In the six-block environment (E-II), among the mid-row agents, Agent 3 primarily uses CS1, and Agent 4 uses CS2.
\item \texttt{stop\_charging:} Battery is the dominant feature, with higher battery levels promoting the action. For the four-block environment (E-I), the battery levels of nearby agents and the queue length are influential. For the six-block environment (E-II), the service mode of neighboring agents also contribute to the decision, reflecting congestion at charging stations.
\item \texttt{go\_to\_depot.} Capacity, distance to depot and distance to next order are the most influential features, indicating that the agents consider the travel cost.
\end{itemize}

\subsubsection{Decision Rule Analysis}

\input{tables/8_s7_decision_rules_EI}
\input{tables/9_s7_decision_rules_EII}

Tables~\ref{tab:decision_rules_EI} and~\ref{tab:decision_rules_6blk} summarize the mean feature values associated with each action across 10 test episodes in the E-I ($\mu = 0.6$) and E-II ($\mu = 0.9$) environments, respectively. Based on these observed mean values, we analyze the operational rationale behind each learned action below.

\texttt{go\_pick}: Agents typically initiate order picking when battery levels are moderate and carrying capacity is approximately half of the maximum. In E-II the gradual increase in battery level during order picking from bottom-row to top-row agents (A1: 61\%, A6: 67\%) reflects the longer travel distances to the
depot for top-row agents, prompting battery conservation.

\texttt{go\_to\_cs}: Agents initiate charging at battery levels of approximately 24\%--28\%, thereby preventing mid-travel battery depletion, and preferentially select the nearest charging station. In Environment E-II, the mid-row agents (Agents 3 and 4) exhibit queue-aware switching between both charging stations. Furthermore, agents appear to coordinate their charging trips by initiating charging when the horizontally adjacent agent has a high battery level, thereby reducing congestion at charging stations.

\texttt{stop\_charging}: In E-I, a spatial asymmetry in charging termination thresholds is observed. Agents 1 and 2 that are located near CS1 and closer to the depot, charge to higher battery levels, whereas Agents 3 and 4 located further away terminate charging earlier. Since Agents 3 and 4 travel longer distances during each order cycle, they trade off charging duration against order throughput more aggressively. In E-II, Agents typically terminate charging after the battery reaches approximately 100\%. In both E-I and E-II, when agents terminate charging, the battery level of neighboring agents emerges as an important factor. In E-I environment, agents consistently terminate charging when their horizontally adjacent partner approaches a low-battery state. Specifically, Agent 1 stops when Agent 2's battery is 37.5\% and Agent 2 stops when Agent 1's is 48.3\%. Similarly, Agent 3 stops when Agent 4's is 43.0\%; Agent 4 stops when Agent 3's is 36.3\%. In E-II environment, coordination extends beyond horizontally adjacent agents to those that predominantly share the same charging station. Agent 1's decision is influenced by Agent 2's low battery, Agent 2's by Agent 3's low battery, and Agent 3's by Agent 1's low battery. Similarly Agent 4 is influenced by Agent 5's low battery, Agent 5 by Agent 6's low battery, and Agent 6 by Agent 4's low battery as shown in Table~\ref{tab:decision_rules_6blk}. This behavior is consistent with the SHAP analysis, which showed that in E-II, Agents 1--3 primarily use CS1, whereas Agents 4--6 primarily use CS2. These findings suggest that agents coordinate their charging decisions to accommodate neighboring agents approaching low-battery states.

\texttt{go\_to\_depot}: Agents go to the depot when carrying capacity is nearly exhausted, maximizing the items delivered per depot trip. The states of neighboring agents fail to exhibit a consistent pattern, indicating that depot decisions are driven primarily by the agent's own carrying capacity.

%% file: tables/8_s7_decision_rules_EI.tex
\begin{table*}[!ht]
\centering
\caption{Mean of state features at the moment each action is chosen.
Battery and capacity shown in absolute units. Queue lengths are number of
agents waiting. Partner battery shown for the most relevant block neighbor.}
\label{tab:decision_rules_EI}
\begin{tabular}{llrrrrr}
\toprule
\textbf{Agent} & \textbf{Action} &
\textbf{Battery} & \textbf{Capacity} &
\textbf{Queue CS1} & \textbf{Queue CS2} &
\textbf{Partner battery} \\
\midrule
\multirow{4}{*}{Agent 1}
 & go\_pick        & 62.6 & 6.1 & 0.5 & 0.6 & A2: 63.2 \\
 & go\_to\_cs1     & 25.4 & 7.2 & 0.0 & 0.6 & A2: 74.2 \\
 & stop\_charging  & 99.0 & 7.2 & 1.1 & 0.7 & A2: 37.5 \\
 & go\_to\_depot   & 57.9 & 1.4 & 0.5 & 0.6 & A2: 65.4 \\
\midrule
\multirow{4}{*}{Agent 2}
 & go\_pick        & 64.2 & 5.8 & 0.5 & 0.6 & A1: 55.4 \\
 & go\_to\_cs1     & 25.9 & 6.6 & 0.1 & 0.7 & A1: 83.0 \\
 & stop\_charging  & 99.9 & 6.6 & 1.0 & 0.6 & A1: 48.3 \\
 & go\_to\_depot   & 62.8 & 0.7 & 0.5 & 0.6 & A1: 55.5 \\
\midrule
\multirow{4}{*}{Agent 3}
 & go\_pick        & 58.7 & 5.7 & 0.6 & 0.4 & A4: 49.3 \\
 & go\_to\_cs2     & 26.1 & 6.6 & 0.6 & 0.2 & A4: 68.8 \\
 & stop\_charging  & 81.9 & 6.6 & 0.6 & 1.1 & A4: 43.0 \\
 & go\_to\_depot   & 58.9 & 0.4 & 0.6 & 0.5 & A4: 47.9 \\
\midrule
\multirow{4}{*}{Agent 4}
 & go\_pick        & 60.1 & 5.6 & 0.6 & 0.4 & A3: 49.5 \\
 & go\_to\_cs2     & 25.2 & 6.9 & 0.6 & 0.1 & A3: 64.0 \\
 & stop\_charging  & 85.3 & 6.8 & 0.6 & 1.1 & A3: 36.3 \\
 & go\_to\_depot   & 60.7 & 0.2 & 0.6 & 0.6 & A3: 52.3 \\
\bottomrule
\end{tabular}
\end{table*}

%% file: tables/9_s7_decision_rules_EII.tex
\begin{table*}[!ht]
\centering
\caption{Mean of state features at the moment each action is chosen in the
six-block environment. Battery and capacity in absolute units.
Queue lengths are number of agents waiting.
Partner battery shown for the most relevant block neighbor.}
\label{tab:decision_rules_6blk}
\begin{tabular}{llrrrrr}
\toprule
\textbf{Agent} & \textbf{Action} &
\textbf{Battery} & \textbf{Capacity} &
\textbf{Queue CS1} & \textbf{Queue CS2} &
\textbf{Partner battery} \\
\midrule
\multirow{4}{*}{Agent 1}
 & go\_pick        & 60.9 & 6.3 & 1.0 & 1.1 & A2: 64.5 \\
 & go\_to\_cs1     & 23.7 & 6.2 & 0.7 & 1.1 & A2: 65.7 \\
 & stop\_charging  & 100.0 & 6.2 & 1.6 & 1.1 & A2: 29.1 \\
 & go\_to\_depot   & 60.1 & 1.9 & 0.9 & 1.1 & A2: 65.5 \\
\midrule
\multirow{4}{*}{Agent 2}
 & go\_pick        & 60.2 & 5.8 & 0.9 & 1.1 & A1: 49.4 \\
 & go\_to\_cs1     & 24.5 & 4.7 & 0.9 & 1.1 & A1: 78.8 \\
 & stop\_charging  & 100.0 & 4.7 & 1.4 & 1.1 & A3: 35.1\\
 & go\_to\_depot   & 67.8 & 0.7 & 0.9 & 1.1 & A1: 46.8 \\
\midrule
\multirow{5}{*}{Agent 3}
 & go\_pick        & 57.7 & 5.8 & 1.0 & 1.1 & A4: 57.1 \\
 & go\_to\_cs1     & 25.7 & 5.0 & 0.8 & 1.1 & A4: 58.9 \\
 & go\_to\_cs2     & 31.1 & 3.4 & 1.1 & 0.9 & A4: 64.1 \\
 & stop\_charging  & 100.0 & 4.7 & 1.4 & 1.2 & A1: 35.6 \\
 & go\_to\_depot   & 69.5 & 0.7 & 1.0 & 1.1 & A4: 56.1 \\
\midrule
\multirow{5}{*}{Agent 4}
 & go\_pick        & 62.4 & 5.6 & 1.0 & 1.1 & A3: 55.5 \\
 & go\_to\_cs2     & 28.0 & 6.2 & 1.1 & 0.9 & A3: 57.9 \\
 & go\_to\_cs1     & 27.2 & 5.9 & 0.8 & 1.3 & A3: 66.9 \\
 & stop\_charging  & 100.0 & 6.1 & 1.1 & 1.6 & A5: 38.4 \\
 & go\_to\_depot   & 67.4 & 0.4 & 1.0 & 1.1 & A3: 56.5 \\
\midrule
\multirow{4}{*}{Agent 5}
 & go\_pick        & 65.1 & 5.6 & 1.0 & 1.0 & A6: 55.8 \\
 & go\_to\_cs2     & 25.8 & 5.9 & 1.0 & 0.9 & A6: 69.7 \\
 & stop\_charging  & 99.8 & 5.9 & 1.0 & 1.6 & A6: 36.3 \\
 & go\_to\_depot   & 76.9 & 0.2 & 1.0 & 1.0 & A6: 55.9 \\
\midrule
\multirow{4}{*}{Agent 6}
 & go\_pick        & 67.2 & 5.6 & 1.0 & 1.0 & A5: 52.8 \\
 & go\_to\_cs2     & 26.3 & 6.4 & 1.0 & 0.9 & A5: 73.8 \\
 & stop\_charging  & 99.9 & 6.4 & 1.1 & 1.5 & A4: 37.4 \\
 & go\_to\_depot   & 75.9 & 0.2 & 1.0 & 1.1 & A5: 47.8 \\
\bottomrule
\end{tabular}
\end{table*}

%% file: sections/9_sensitivity.tex
\section{Sensitivity}
Real warehouses are highly dynamic, and to implement a DRL framework, it is necessary to ensure the model's robustness. Hence, we performed additional experiments under different settings. 

\subsection{Sensitivity to training and testing durations} 

\begin{figure}[!ht]
    \centering
    \begin{subfigure}{0.45\textwidth}
        \centering
        \includegraphics[width=\linewidth]{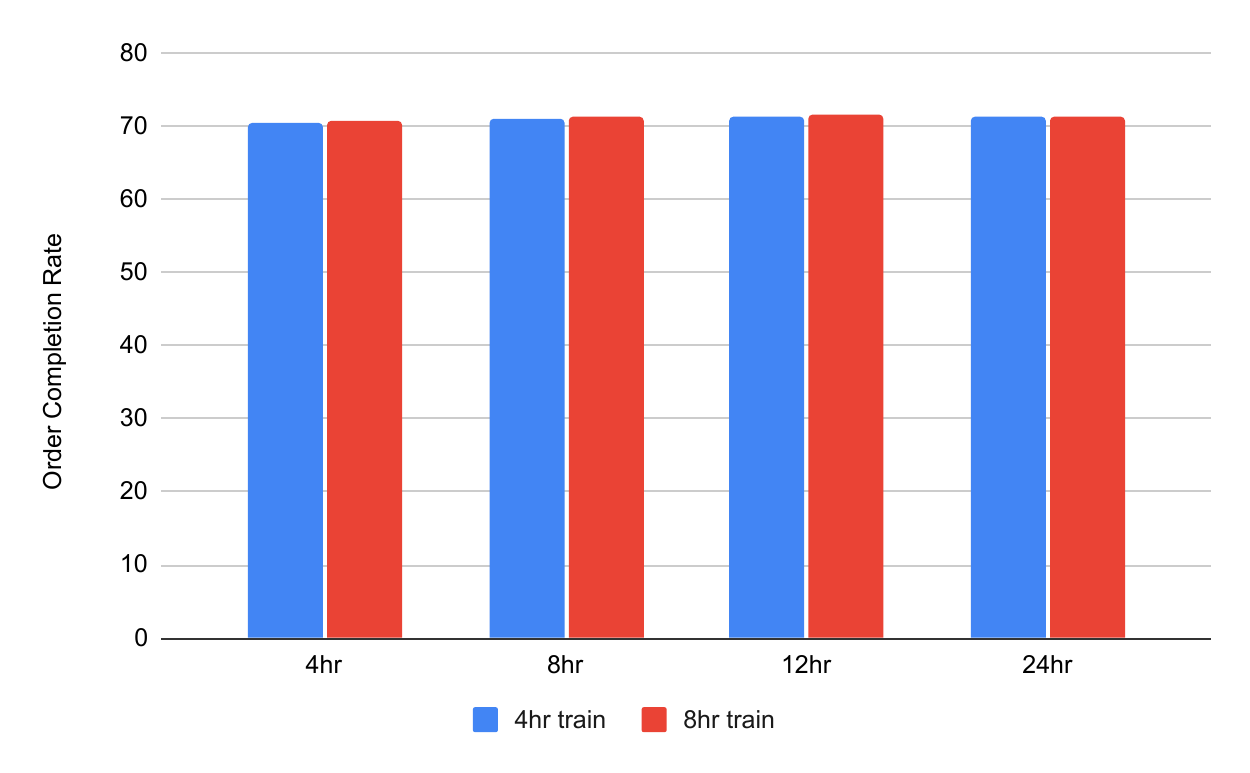}
    \end{subfigure}
    \caption{Order completion rate at different training and testing time in E-II environment}
    \label{fig:s9_1}
\end{figure}

To evaluate whether training duration significantly affects testing performance, models were trained using 4-hour and 8-hour simulations in the E-II environment (with $\mu = 0.75$) and subsequently evaluated over simulation horizons of 4, 8, 12, and 24 hours. As illustrated in Figure~\ref{fig:s9_1}, all configurations achieve comparable total order completion rates. This behavior can be attributed to the repetitive and stationary nature of warehouse operations, where agents experience similar battery utilization and consistent travel patterns regardless of the simulation horizon. Consequently, no significant gain in performance metrics is achievable by training over a longer time period.

\subsection{Dynamic order arrival rate}
\begin{figure*}[!ht]
    \centering
    \begin{subfigure}{0.45\textwidth}
        \centering
        \includegraphics[width=\linewidth]{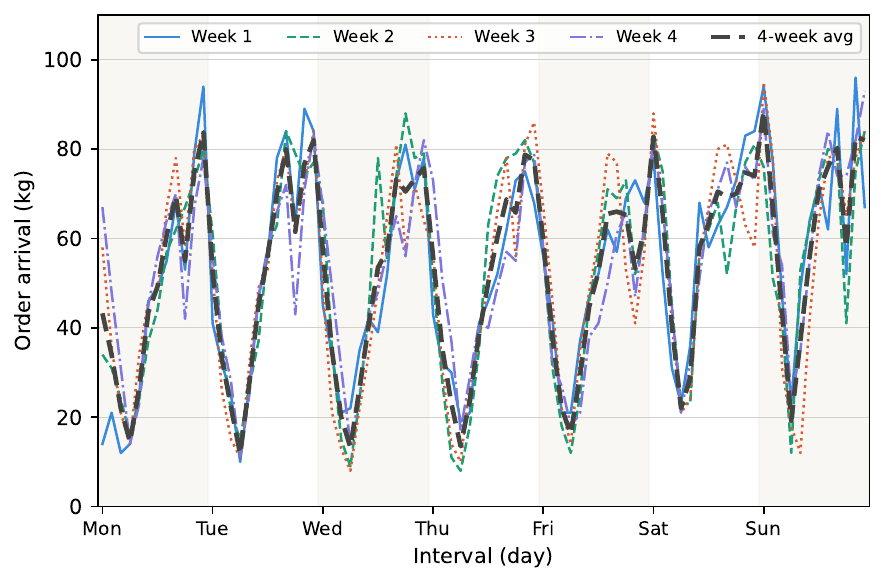}
        \caption{Arrival volume in kg for a 4-week period in source data}
    \end{subfigure}%
    \hspace{0.01\textwidth}
    \begin{subfigure}{0.45\textwidth}
        \centering
        \includegraphics[width=\linewidth]{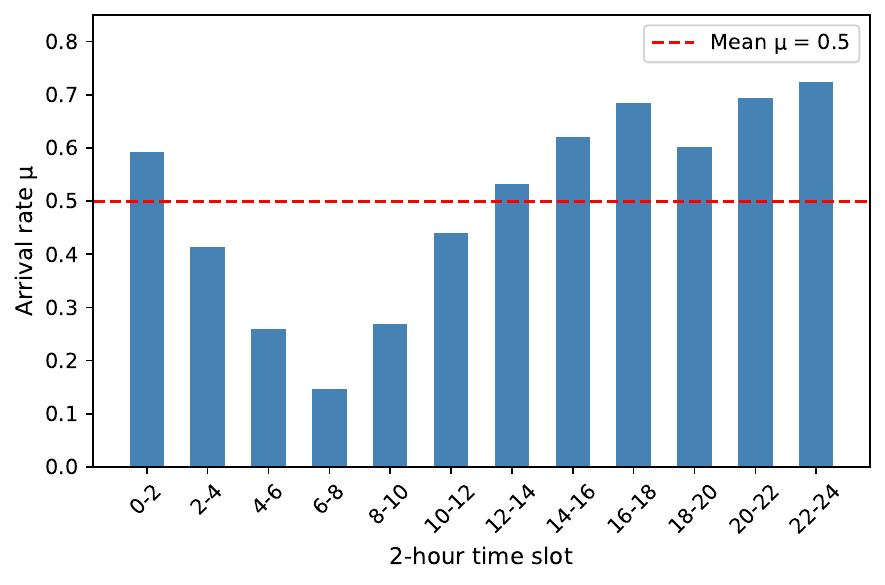}
        \caption{Derived arrival rate $\mu$ for a 24 hour period}
    \end{subfigure}%
    \caption{Dynamic arrival rate}
    \label{fig:s9_2}
\end{figure*}

In practice, orders rarely arrive at a constant rate. To simulate this real-world scenario, the model was trained on dynamic order arrival rate. The order arrival process follows a piecewise-constant Poisson Process, with arrival rates derived from empirical B2C e-commerce order data collected in 2-hour intervals from a real distribution center \citep{arleung2019}. The source data reports order volume in kg for every 2-hour slot over a four-week period (all seven days), as shown in Figure~\ref{fig:s9_2}. To derive a time-varying arrival schedule, the mean arrival volume in kg per 2-hour slot was computed by averaging all 28 daily observations (4 weeks $\times$ 7 days). Each slot mean is then expressed as a ratio relative to the overall daily mean, yielding 12 relative weights that capture the intra-day arrival pattern. Since the source data reports order volume in kg rather than order count, we apply these relative weights directly to our simulation mean of $\mu= 0.5$ orders/timestep, producing a piecewise-constant arrival schedule as shown in Table~\ref{tab:arrival_profile}.

\begin{table}[!ht]
\centering
\caption{Intra-day order arrival profile}
\label{tab:arrival_profile}
\renewcommand{\arraystretch}{1.15}
\begin{tabular}{clccc}
\toprule
\textbf{Slot} &
\textbf{Period (h)} &
\textbf{Avg volume (kg)} &
\textbf{Ratio $r_i$} &
\textbf{$\mu_i$ (orders/s)} \\
\midrule
0  & 00:00--02:00 & 62.8 & 1.189 & 0.5946 \\
1  & 02:00--04:00 & 43.9 & 0.832 & 0.4158 \\
2  & 04:00--06:00 & 27.6 & 0.523 & 0.2613 \\
3  & 06:00--08:00 & 15.8 & 0.299 & 0.1494 \\
4  & 08:00--10:00 & 28.6 & 0.541 & 0.2704 \\
5  & 10:00--12:00 & 46.6 & 0.883 & 0.4415 \\
6  & 12:00--14:00 & 56.5 & 1.070 & 0.5348 \\
7  & 14:00--16:00 & 65.8 & 1.245 & 0.6223 \\
8  & 16:00--18:00 & 72.4 & 1.371 & 0.6855 \\
9  & 18:00--20:00 & 63.7 & 1.206 & 0.6030 \\
10 & 20:00--22:00 & 73.5 & 1.391 & 0.6953 \\
11 & 22:00--24:00 & 76.7 & 1.452 & 0.7261 \\
\midrule
   & \textbf{Daily mean} & \textbf{52.8} & \textbf{1.000} & \textbf{0.5000} \\
\bottomrule
\end{tabular}
\end{table}

\begin{figure}[!ht]
    \centering
    \includegraphics[width=0.45\linewidth]{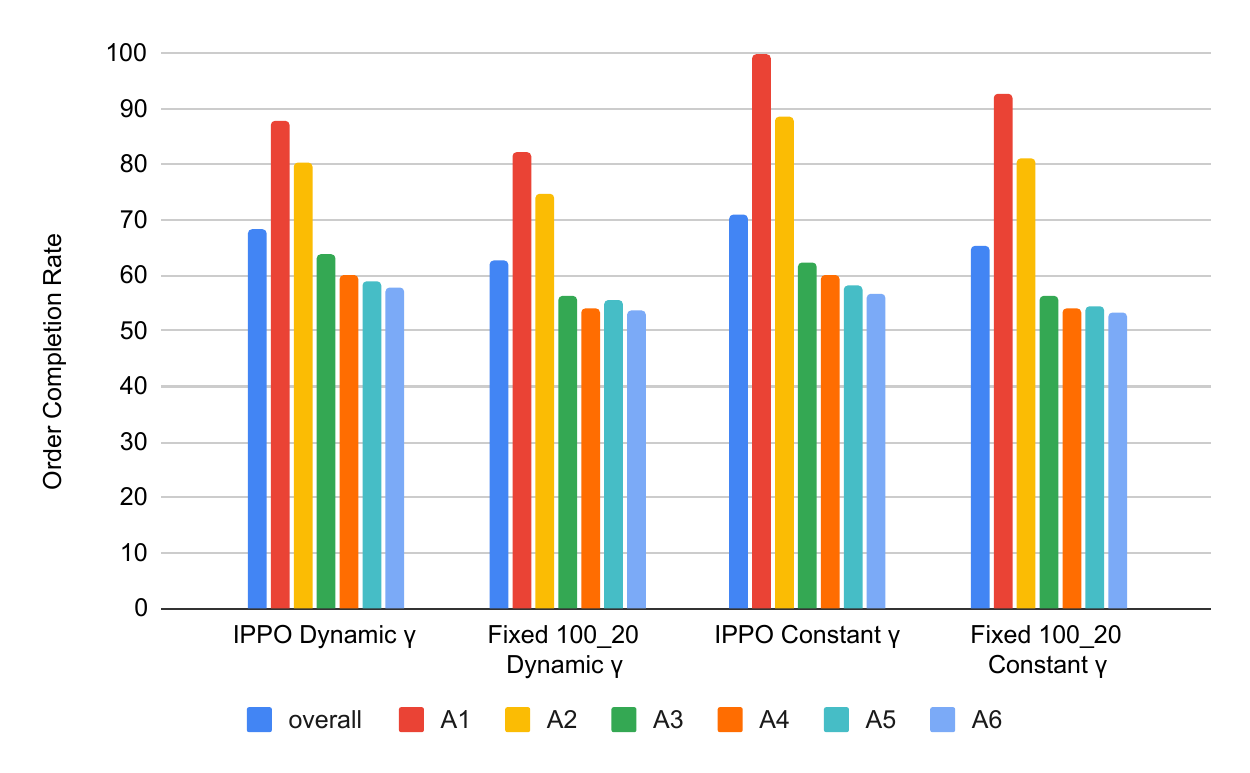}
    \caption{Effect of dynamic arrival rate}
    \label{fig:s9_3}
\end{figure}

To evaluate the robustness of the proposed IPPO framework, the dynamic arrival schedule was applied to the IPPO model in the E-II environment with $\mu =$ 0.75 and to the best performing benchmark \textit{Fixed 100\_20}. As shown in Figure~\ref{fig:s9_3}, IPPO continues to outperform the benchmark, exhibiting only a marginal drop in order completion rate compared with the constant arrival rate scenario.